%% file: main.tex
\documentclass[11pt,a4paper]{article}
\usepackage{times,latexsym}
\usepackage{url}
\usepackage[T1]{fontenc}
\usepackage{booktabs,amsmath}
\usepackage{todonotes}
\usepackage{rotating,slashbox}

\usepackage[]{tacl2018v2}

\usepackage{xspace,mfirstuc,tabulary,graphicx}

\newcommand{\wmt}[1]{\textsc{WMT-{#1}}\xspace}
\newcommand{\wsmt}{\textsc{WMT}\xspace}

\newcommand{\M}{\hphantom{$-$}}

\newif\iftaclinstructions
\taclinstructionsfalse 
\iftaclinstructions
\renewcommand{\confidential}{}
\renewcommand{\anonsubtext}{(No author info supplied here, for consistency with
TACL-submission anonymization requirements)}
\newcommand{\instr}
\fi

\iftaclpubformat 

\else

\fi

\newcommand{\phicomment}[1]{\textcolor{blue}{PHI: #1}}
\newcommand{\onesig}[1][]{$^{*}$\hphantom{#1}}

\title{Translationese in Machine Translation Evaluation}

\taclpubformattrue
\author{Yvette Graham \\
ADAPT Research Centre\\
Dublin City University\\
\tt{ \small yvette.graham@gdcu.ie}
\And
Barry Haddow\\
School of Informatics \\ University of Edinburgh \\ \tt{\small bhaddow@inf.ed.ac.uk}
\And
Philipp Koehn\\
Dept of Computer Science \\ Johns Hopkins University \\ \tt{\small phi@jhu.edu}
}

\date{}

\begin{document}
\maketitle

\input{sec-00-abstract}
\input{sec-01-intro}
\input{sec-02-relwork}

\input{sec-04-exp}

\input{sec-05-reeval}

\input{sec-06-conc}

\bibliography{tacl2018}
\bibliographystyle{acl_natbib}

\end{document}

%% file: sec-00-abstract.tex
\begin{abstract}
  The term \emph{translationese} has been used to describe the presence of unusual features in translated text. In this paper, we provide a detailed analysis of the adverse effects of translationese on machine translation evaluation results. Our analysis shows evidence to support differences in text originally written in a given language relative to translated text and this can potentially negatively impact the accuracy of machine translation evaluations. For this reason we recommend that reverse-created test data be omitted from future machine translation test sets. 
  
  In addition, we provide a re-evaluation of a past high-profile machine translation evaluation claiming human-parity of MT, as well as analysis of the since re-evaluations of it. We find potential ways of improving the reliability of all three past evaluations. One important issue not previously considered is the statistical power of significance tests applied in past evaluations that aim to investigate human-parity of MT. Since the very aim of such evaluations is to reveal legitimate ties between human and MT systems, power analysis is of particular importance, where low power could result in claims of human parity that in fact simply correspond to Type II error. We therefore provide a detailed power analysis of tests used in such evaluations to provide an indication of a suitable minimum sample size of translations for such studies.
  
  Subsequently, since no past evaluation that aimed to investigate claims of human parity ticks all boxes in terms of accuracy and reliability, we rerun the evaluation of the systems claiming human parity.   Finally, we provide a comprehensive checklist for future machine translation evaluation.
  
\end{abstract}

%% file: sec-01-intro.tex
\section{Introduction}\label{intro}

Human-translated text is thought to display features that deviate to some degree from those of text originally composed in the that language. \newcite{Bakeretal:93} report that translated text can: be more explicit than the original source, less ambiguous, simplified (lexical, syntactically and stylistically); display a preference for conventional grammaticality; avoid repetition; exaggerate target language features; as well as display features of the source language. 
The term \emph{translationese} has often been used to describe the presence of such phenomenon in translated text.  

\begin{figure}
\begin{center}
\includegraphics[width=0.45\textwidth]{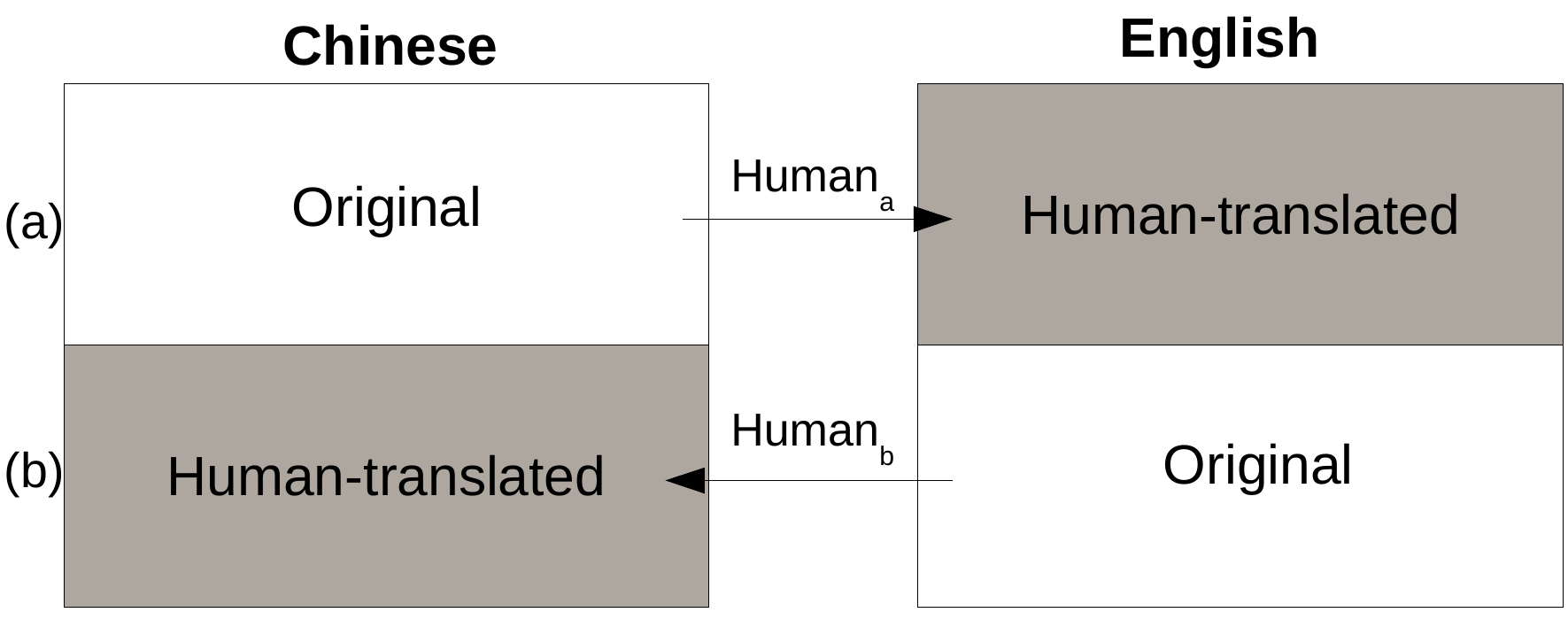}
\caption{Creation of MT test sets for machine translation testing of 
Chinese to English, with potential translationese pollution highlighted in gray.}\label{translationese-pic}
\end{center}
\end{figure}

Standard evaluation protocol in Machine Translation (MT) comprises system tests on a sample of human-translated text. Since creating this human-translated text is expensive, re-use of test sets for both directions of translation is commonplace, without
regard to whether source or target contain features of translationese.  For example, translation shared tasks at the Conference on Machine Translation ({\wsmt}) \cite{wmt18} generally test translation between a given language pair as depicted in 
Figure \ref{translationese-pic} for testing Chinese to English translation. Portion (a) of the test data 
(accounting for approximately 50\% of sentences) is made up of text that originated in Chinese and was 
human-translated into English, while portion (b) (i.e. the remaining 50\%), was translated in the opposite direction, originating in English with manual translation into Chinese.
The motivation for creating the test data in this way is to create test sets for both directions simultaneously (so at no extra cost).

Although translationese has been cited as a likely confound in MT evaluation results in the past \cite{Toraletal:18,laeublietal:18}, to the best of our knowledge, no detailed investigation into the impact of translationese on the accuracy of MT evaluation has been reported to date. With this aim, we examine the degree to which translationese phenomena may impact human and automatic evaluation results in MT.
We firstly examine past results of  {\wsmt} shared tasks, a main venue for MT evaluation, and reveal that although system rankings are overall very similar for human evaluation of forward and reverse test data, in a small number of cases system rankings diverge to a more serious degree.
For example, for Turkish-English translation at WMT-18 forward and reverse system rankings correlate at only $r$ = 0.703 in one case.
Besides human evaluation, much more concerning is the divergence in forward and reverse rankings when BLEU is relied upon for evaluation of systems, where the correlation can be as low as 0.106 in the worst case.

Subsequently, we provide a reassessment of a human evaluation previously criticized for including reverse-created test data that claimed human parity of Chinese to English MT.
We reveal insights into additional potential sources of inaccuracy of conclusions beyond the presence of translationese
with the aim of preventing future inaccuracies. To this end, we provide a concise and clear checklist of considerations that should be taken into account when planning or reviewing MT evaluations.

%% file: sec-02-relwork.tex
\section{Related Work}\label{relwork}


\newcite{hassanetal:18} provide one of the earliest claims in MT of systems achieving human-parity in terms of the quality of translations. 
\newcite{laeublietal:18} and \newcite{Toraletal:18} both question the
reliability of conclusions due to it following the 50/50 set-up  of test data creation (shown in Figure 
\ref{translationese-pic}), highlighting the inclusion of reverse-created test data as a likely confound. 
\newcite{laeublietal:18} and \newcite{Toraletal:18} repeat the human evaluation of translations produced by \newcite{hassanetal:18} only for test data that originated in the source language and with some additional distinctions.

Firstly, and making a positive change, both \newcite{laeublietal:18} and  \newcite{Toraletal:18} include more context than the original sentence-level evaluation, the former now asking human judges to assess entire documents, and the latter involving assessment of MT output sentences in the order that they appeared in original documents. Secondly, both reassessments again move away from the evaluation method employed in \newcite{hassanetal:18}, Direct Assessment \cite{Grahametal:16}, and revert to an older method of human evaluation, relative ranking, no longer used at WMT for evaluation of systems.

In addition, in both re-evaluations, besides use of older evaluation methodologies, another  concern is that they were limited to only a small number of human judges with low levels of inter-annotator agreement. Therefore, although both re-evaluations improved the methodology employed in two respects, by eliminating reverse-created test data and including more context, both potentially include other sources of inaccuracy, such as lack of reliability of human judges when human evaluation takes the form of relative ranking \cite{WMT:07,WMT:08,WMT:09,WMT:10,WMT:11,WMT:12,WMT:13,WMT:14,WMT:15,WMT:16}. 

Furthermore, \newcite{Toraletal:18} employ Trueskill to reach the  conclusion that the MT system in question has not achieved human performance, and although Trueskill has been used in past WMT evaluations to produce system rankings, its aim is to minimize the number of judgments required to produce those rankings when resources are limited. Results may not be directly comparable with results of standard statistical significance tests
 therefore, now current practice at {\wsmt} evaluations. 

Finally, neither \newcite{Toraletal:18} nor \newcite{laeublietal:18} discuss statistical power of significance tests used to distinguish the performance of system and human, an important aspect of evaluation and one of particular importance with respect to evaluations that aim to investigate claims of human parity, where Type II error could result in false claims.

Besides criticisms already made of the human evaluation
in \newcite{hassanetal:18}, an additional aspect of importance not yet highlighted is the proportion of distinct translations that were included in the original human-parity evaluation of systems, a consideration that also relates strongly to the question of statistical power. 
In most MT human evaluations, it is not feasible to evaluate the full test set of sentences for all systems and it is common to instead evaluate a \emph{sample of translations}, usually drawn at random from the test data.
In current {\wsmt} evaluations, for example, translations of all test sentences produced by all participating systems are pooled and a random sample are human-evaluated. This method ensures that as great a number as possible of \emph{distinct test sentences} are examined.
Alongside system performance estimates, {\wsmt} also reports the number of distinct test sentences evaluated, $n$, and it is this number that they consider the \emph{sample size} used for statistical significance tests subsequently used to draw conclusions about which competing systems outperform others. For example, all else being equal, a difference in system performance estimates for a pair of systems computed from a \emph{larger set of distinct translations} is interpreted as \emph{more reliable}.

Other MT human evaluations, despite claims of following {\wsmt} human evaluation methodology, have diverged from this method of sample size computation, however, including the human-parity evaluation of \newcite{hassanetal:18} and \newcite{laeublietal:18}.
For example, although a large  sample of human judgments is reported as $n$ $\geq$ 1,827 per system in \newcite{hassanetal:18}, firstly this number in fact included quality control check translations, generally removed from data before computing sample sizes. More importantly however, very high numbers of repeat evaluations of the same translations were included in the human-parity evaluation of \newcite{hassanetal:18}. In other words, a very low number of distinct test sentences were in fact  human evaluated despite reporting a large sample size.
The method of computing sample size therefore diverges from that reported of {\wsmt} evaluations in a small but important way. The sample size reported instead corresponds to the total number of human ratings collected as opposed to distinct test sentences (as in {\wsmt} evaluations). In this current work, we make this important distinction explicit by referring to the number of distinct test sentences evaluated as $n$ and the number of human judgments collected as $N$. We also recommend this distinction be made and adopted as common practice in future human evaluations of MT. 

Table \ref{hassan-orig} shows results reproduced from the \newcite{hassanetal:18} data set, where we now report both the number of human judgments collected, $N$, and the number of distinct test sentences included, $n$, in addition to adding separate results for forward and reverse-created test data. Only when tested on the less legitimate reverse direction data does MT now appear to outperform human translation.
Nonetheless, when interpreting results in Table \ref{hassan-orig}, it is important to remember, however,  that the reliability of even the conclusions drawn from forward-created test data only is still uncertain however, due to the small $n$, as only 92 distinct translations were in fact included in the evaluation claiming human parity. It remains a possibility that, for example, had the number of distinct test sentences evaluated been higher  that distinct conclusions would also be drawn.

Since the original human evaluation in \newcite{hassanetal:18} was hampered by low numbers of distinct test sentences and both subsequent re-evaluations hampered by somewhat out-dated human evaluation methodologies and low inter-annotator agreement levels between human judges, we rerun  the evaluation using the original translation data included in \newcite{hassanetal:18} with entirely up-to-date {\wsmt} human evaluation methodology in addition to ensuring that a sufficiently large sample of distinct translations are assessed by human judges. We also take into account the very legitimate criticism made by both \newcite{Toraletal:18} and \newcite{laeublietal:18} and include document-level context in the human evaluation. Furthermore, since no previous evaluation has included statistical power analysis, prior to running our own human evaluation, we examine the power of significance tests to estimate a suitable sample size to decrease the likelihood of Type II error leading to conclusions of human parity due to the application of a low powered test.

Prior to rerunning the evaluation, we examine potential issues for MT evaluation when test data created in the reverse direction to testing. Despite being identified by \newcite{Toraletal:18} and \newcite{laeublietal:18} as a serious cause of concern in MT evaluations, to the best of our knowledge no previous study exists that examines in detail the degree to which reverse-created test data may have skewed past results. The sections that follow therefore include an investigation into the issue of translationese in MT evaluation, in addition to a re-evaluation of \newcite{hassanetal:18} data with all potential sources of criticism, in terms of test data and evaluation methodology, now taken into account and corrected. 

In other work, past MT evaluations have investigated the effect of using translated and original data for training statistical machine translation (SMT) systems \cite{lamb2012trans}, revealing that training data created via translation, as opposed to data sourced from text originally written in a given language, achieves better results
for systems in some cases.

\begin{table}
    \centering
    \scriptsize
    \input{tbl-04-hassan-orig.tex}
    \caption{Results of \newcite{hassanetal:18} for forward, reverse and both test set creation directions reproduced from published data set.
    $N$ is the number of human judgments collected for that system while $n$ is the number of distinct translations assessed for that system,
    Reference-HT are human translations created by \cite{hassanetal:18},
    Reference-PE are the outputs of an online MT system after human correction, Reference-WMT are the original WMT reference translations.}
    \label{hassan-orig}
\end{table}


\begin{table*}[]
    \centering
    \scriptsize
    \input{tbl-07-hassan-effectsize}
    \caption{Effect size for all systems included in \newcite{hassanetal:18}}
    \label{hassan-effect}
\end{table*}

%% file: tbl-04-hassan-orig.tex
\begin{tabular}{ccc}

\begin{tabular}{ccccc}
\toprule
\multicolumn{5}{c}{FWD}\\
Ave. & z & $n$ & $N$ & System \\
\midrule
67.1 & \M0.185    & 92 & 828 & Reference-HT \\
64.8 & \M0.048    & 92 & 828 & Combo-5 \\
64.3 & \M0.042    & 92 & 828 & Combo-6 \\
64.3 & \M0.023    & 92 & 828 & Combo-4 \\
64.1 & \M0.020    & 92 & 828 & Reference-PE \\
61.1 & $-$0.144 & 92 & 828 & Reference-WMT \\ \midrule
56.2 & $-$0.345 & 92 & 828 & Sogou \\         \midrule
50.9 & $-$0.580 & 92 & 828 & Online-A-1710 \\
48.5 & $-$0.717 & 92 & 828 & Online-B-1710 \\

\end{tabular} \\

\begin{tabular}{ccccc}
\toprule
\multicolumn{5}{c}{REV}\\
Ave. & z & $n$ & $N$ & System \\
\midrule
73.8 & \M0.434    & 89 & 801 & Combo-6 \\
73.2 & \M0.393    & 89 & 802 & Combo-5 \\
72.8 & \M0.392    & 89 & 801 & Combo-4 \\ \midrule
70.3 & \M0.256    & 89 & 801 & Reference-PE \\
70.0 & \M0.252    & 89 & 801 & Reference-HT \\
68.8 & \M0.167    & 89 & 801 & Sogou \\ \midrule
63.0 & $-$0.089 & 89 & 801 & Reference-WMT \\
60.0 & $-$0.214 & 89 & 801 & Online-B-1710 \\
61.1 & $-$0.217 & 89 & 802 & Online-A-1710 \\

\end{tabular} \\

\begin{tabular}{ccccc}
\toprule
\multicolumn{5}{c}{BOTH}\\
Ave. & z & $n$ & $N$ & System \\
\midrule
69.0 & \M0.235  & 181 & 1,629 & Combo-6 \\
68.5 & \M0.218  & 181 & 1,629 & Reference-HT \\
68.9 & \M0.218  & 181 & 1,630 & Combo-5 \\
68.5 & \M0.204  & 181 & 1,629 & Combo-4 \\
67.1 & \M0.136  & 181 & 1,629 & Reference-PE \\ \midrule
62.4 & $-$0.093 & 181 & 1,629 & Sogou \\
62.0 & $-$0.117 & 181 & 1,629 & Reference-WMT \\ \midrule
55.9 & $-$0.402 & 181 & 1,630 & Online-A-1710 \\
54.1 & $-$0.469 & 181 & 1,629 & Online-B-1710 \\
\bottomrule
\end{tabular} &

\end{tabular}

%% file: tbl-07-hassan-effectsize.tex
\small
\begin{tabular}{lccccccccccccccccccccccc}
\toprule
  & \rotatebox{90}{Ref-HT} & \rotatebox{90}{Combo-5} & \rotatebox{90}{Combo-6} & \rotatebox{90}{Combo-4} & \rotatebox{90}{Ref-PE} & \rotatebox{90}{Ref-WMT} & \rotatebox{90}{Sogou} & \rotatebox{90}{Online-A} & \rotatebox{90}{Online-B}\\
 \midrule
Ref-HT & $-$ & 0.565 & 0.554 & 0.598 & 0.598 & 0.609 & 0.728 & 0.761 & 0.880 \\
Combo-5 & 0.435 & $-$ & 0.174 & 0.293 & 0.533 & 0.587 & 0.783 & 0.793 & 0.848 \\
Combo-6 & 0.446 & 0.239 & $-$ & 0.359 & 0.511 & 0.598 & 0.793 & 0.793 & 0.880 \\
Combo-4 & 0.402 & 0.228 & 0.272 & $-$ & 0.500 & 0.609 & 0.750 & 0.783 & 0.826 \\
Ref-PE & 0.402 & 0.467 & 0.489 & 0.500 & $-$ & 0.554 & 0.750 & 0.793 & 0.880 \\
Ref-WMT & 0.391 & 0.413 & 0.402 & 0.391 & 0.435 & $-$ & 0.598 & 0.696 & 0.793 \\
Sogou & 0.272 & 0.207 & 0.196 & 0.239 & 0.250 & 0.402 & $-$ & 0.663 & 0.728 \\
Online-A & 0.239 & 0.196 & 0.196 & 0.207 & 0.207 & 0.304 & 0.326 & $-$ & 0.533 \\
Online-B & 0.120 & 0.152 & 0.120 & 0.174 & 0.098 & 0.196 & 0.272 & 0.467 & $-$ \\

\bottomrule
\end{tabular}

%% file: sec-04-exp.tex
\section{Translationese}

When testing MT systems it seems more natural
to test systems in the forward direction:
by taking text that genuinely originated in the source language, 
inputting it to a given MT system, and comparing the output
with human translation of the same sentences.
However, as described previously, 
as an artifact of WMT evaluations
being carried out in both translation directions, it is common in MT evaluation for only around 50\% of test sentences to
be created in the forward direction with the remaining 
created in the reverse direction to testing, or even 
select test data without taking into account test data
creation direction.

It is thought however that using reverse-created test data makes the evaluation unrealistically easy \cite{Toraletal:18,laeublietal:18}, because in real-world MT
scenarios, input text is unlikely to very often comprise text that has already been translated from the target language.
Due to the possibility that the 
portion of test data created in the reverse direction
could artificially boost MT evaluation results, we investigate
with past evaluation data 
the degree to which this is actually the case.
We therefore compare results of systems when test data is split according to the creation direction and examine differences in scores for systems in terms of both human and automatic metrics.

\subsection{Human Evaluation}\label{human-eval}

In order to examine differences in human evaluation results for MT systems with respect to the presence of translationese as a possible confound, we firstly examine systems participating in past evaluation campaigns at {\wmt{17}} and {\wmt{18}}, where direct assessment (DA) was employed as the official human evaluation measure.\footnote{Prior to 2017, the method of human evaluation employed at WMT was relative ranking, where a preference between competing pairs of translations was provided by human judges, only recording whether or not the one translation was considered better or worse than the other. This method of human evaluation cannot be used to analyze absolute quality judgments for the reverse and forward test data as we do with DA scores.}
We compute two separate human evaluation scores for each system.
Firstly, for each individual system, we compute its \emph{forward DA score}, comprising the average DA score computed only
for test sentences that were created in the \emph{same direction as testing}. Secondly, a corresponding \emph{reverse DA score} is computed as the average DA score for 
MT output sentences corresponding to test data created in the \emph{opposite direction to testing}.
Then, to examine the extremity to which MT human evaluation results may differ when systems are tested in the reverse as opposed to forward direction, we subtract a given system's forward DA score (expected to be lower than its reverse counterpart) from its reverse DA score (expected to be higher than its forward counterpart). This provides the difference in human DA scores for each system, with positive differences expected in general since reverse-created test data is thought to be an artificially easier test for MT systems.

\begin{figure*}
\begin{center}
\begin{tabular}{cc}
\includegraphics[width=1.0\textwidth]{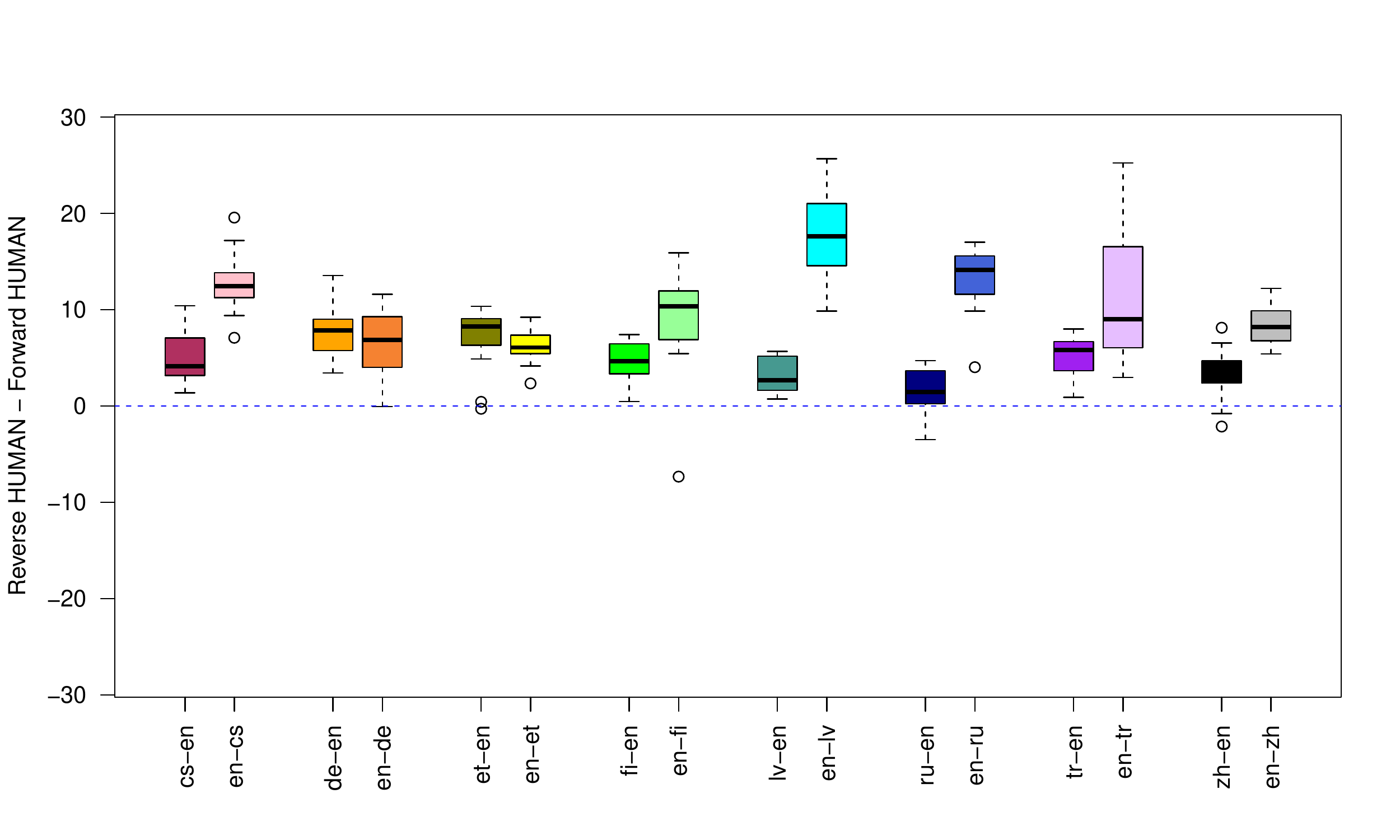} \\
\end{tabular}
\caption{Differences in human evaluation DA scores 
for test sentences created in the reverse direction to
testing and those created in the same/forward
direction to testing broken down by language pair, showing
that reverse human
evaluation scores higher than forward ones in almost all
cases. \phicomment{there is a lot of white space in the bottom half of this chart - maybe cut it off}}\label{diff-by-lp-boxplot}
\end{center}
\end{figure*}

Figure~\ref{diff-by-lp-boxplot} shows the distribution of DA score differences (reverse DA $-$ forward DA) for all systems participating 
in {\wmt{17}} and {\wmt{18}} news translation shared task
broken down by language pair, where positive differences for systems indicate a higher human evaluation score when systems are tested in the reverse direction relative to the corresponding  forward direction DA score.
Table \ref{human-diff-by-lp} shows the mean and standard
deviation of score differences for the same set of systems.
\begin{table}
\begin{center}
\input{tbl-01-humandiff-by-lp}
\caption{Comparison of human evaluation scores
of MT systems participating in {\wmt{17}} and {\wmt{18}}
for test data created in the same/forward direction (F) and reverse (R)
direction, where
R$>$F (\%) = the proportion of systems with a 
reverse DA score greater than its forward score 
for precisely the same test scenario;
R$-$F $\mu$ $=$ mean of the difference in
reverse and forward DA scores;
R$-$F $\sigma$ $=$ standard deviation of the difference in
reverse and forward DA scores;
$n =$ number of MT systems.}
\label{human-diff-by-lp}
\end{center}
\end{table}

As can be seen from the box plot in Figure \ref{diff-by-lp-boxplot}
and results in Table \ref{human-diff-by-lp}, almost all reverse DA scores are higher than equivalent forward DA scores. 
This confirms the suspicion that absolute human evaluation results are in general higher when test data is created in the reverse direction to testing, ranging from the least average difference of 1.65, for Russian to English translation, up to the largest and substantial average 
difference of 18.04 for English-Latvian.

It is important when carrying out such a comparison, however,
to consider the degree to which splitting DA scores in the way we have done here really provides a good and valid comparison.
One thing to consider is human assessors and, more specifically,
was there any difference in human assessors between forward and reverse DA scores? For example, if human evaluation of forward and reverse test sentences were carried out by two different groups of human assessors, this damages the validity of the comparison, since differences in forward and reverse scores could be caused to some unknown degree by differences in human judge scoring strategies as opposed to differences in text. 
Human evaluation at {\wsmt} thankfully includes randomization of test 
sentences that distributes close to equal proportions of forward and reverse test sentences to each human judge however, and this ensures that differences in human judge scoring strategies will not negatively impact the validity of our comparison of forward and reverse scores.

Another worthwhile consideration is how splitting the test data may or may not impact the intended interpretation of DA scores. The fact that DA scores are simply a straightforward average of absolute scores for sentences, however, ensures that splitting human evaluation results for forward and reverse direction testing does not change the interpretation of each separate human score, since both remain a simple average of sentence scores.

A final consideration about the validity of our comparison of forward and reverse DA scores is the fact that splitting the test data does of course result in two distinct sets of test sentences. It is possible therefore that there remains something we have not taken into account about a given set of sentences (besides its creation direction) that could impact the difficulty of translation, such as a more difficult topic in the forward direction as opposed to the reverse direction. However, although the test sentences in the forward and reverse sets are distinct sentences, the fact that both sets are randomly selected news articles helps provide a sufficiently valid comparison. 

In the section that follows, we will compare BLEU scores for the forward and reverse directions, which, as we will see, comprises a less straightforward comparison than human evaluation DA scores.

\subsection{BLEU}

Besides human evaluation, the performance of MT systems is often measured using automatic metrics, the most common of which remains to be the BLEU score \cite{Papineni:2002}, and we therefore compare  forward and reverse BLEU scores for systems participating in past evaluation campaigns. 
Figure \ref{diff-by-lp-boxplot} shows a box plot
of absolute differences in BLEU scores for systems (reverse BLEU $-$ forward BLEU) participating in {\wsmt}
news translation tasks from 2015 to 2018, and Table \ref{bleu-diff-by-lp} 
shows differences in terms of mean and standard deviation,
as well as proportions of systems with a higher reverse than
forward BLEU score for the same set of systems.

\begin{figure*}
\begin{center}
\includegraphics[width=0.85\textwidth]{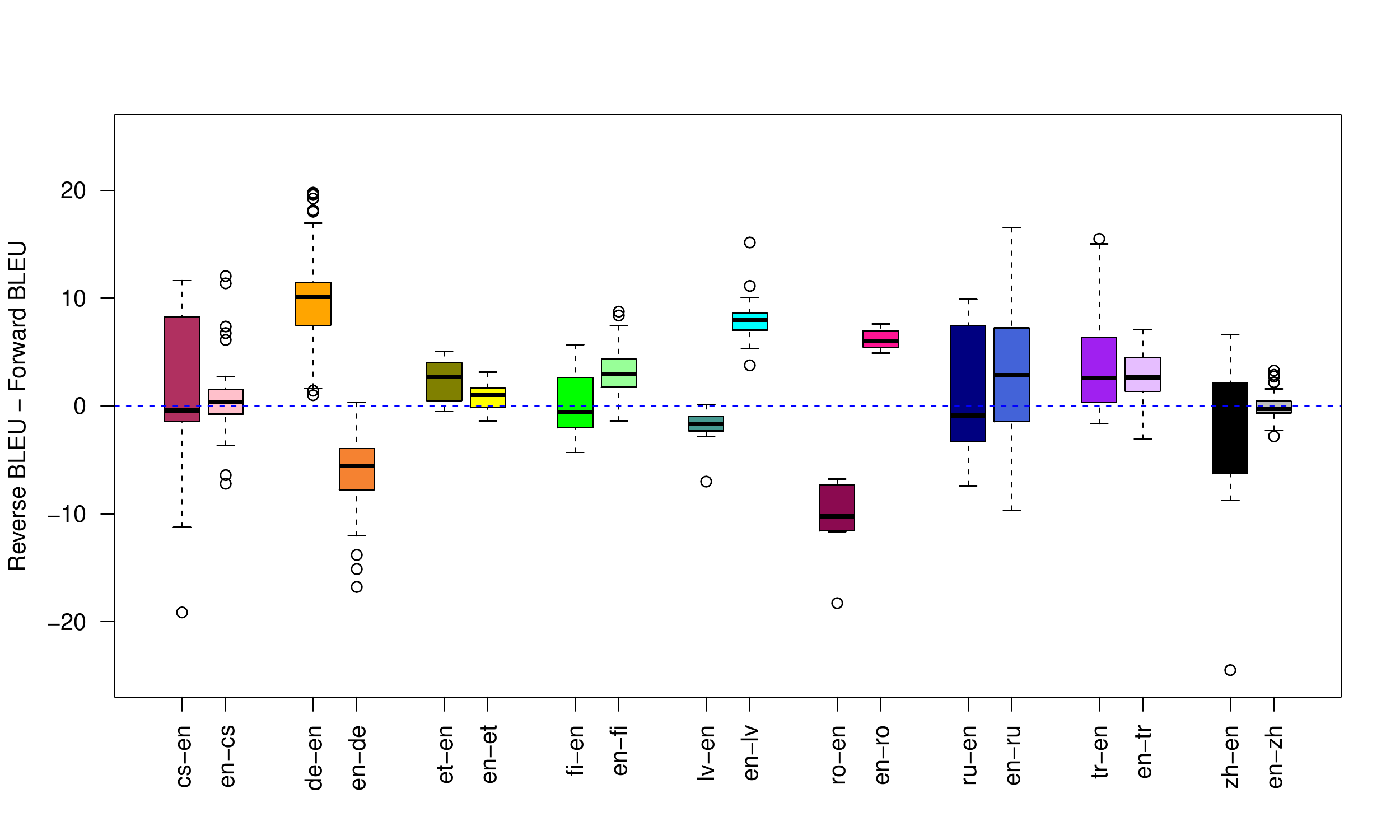}
\caption{Differences in BLEU scores for systems participating
in {\wmt{15}}--{\wmt{18}} news translation task computed
for test sentences created in the reverse direction to
testing and those created in the same/forward
direction to testing broken down by language pair, showing
a mix of positive and negative differences in BLEU scores
depending on test set creation direction.}\label{bleu-diff-by-lp-boxplot}
\end{center}
\end{figure*}

Somewhat surprisingly, results in Figure \ref{bleu-diff-by-lp-boxplot} and
Table \ref{bleu-diff-by-lp} do not display the same trend of higher reverse scores observed in human evaluation results in Section \ref{human-eval}.
Counter expectation there is a clear mix of positive and negative BLEU score differences for several language pairs, as forward BLEU scores are higher than equivalent reverse BLEU scores (cs-en, en-cs, en-de,  en-et, fi-en, lv-en, ro-en, ru-en, en-ru, zh-en and en-zh). 
\begin{table}
\begin{center}
\input{tbl-00-bleudiff-by-lp}
\caption{Comparison of BLEU scores
of MT systems participating in {\wmt{15}} -- {\wmt{18}}
for test data created in the same/forward (F) and reverse (R)
direction, where
R$>$F (\%) = the proportion of systems with a 
reverse BLEU score greater than its forward score 
for precisely the same test scenario;
R$-$F $\mu$ $=$ mean of the difference in
reverse and forward BLEU scores;
R$-$F $\sigma$ $=$ standard deviation of the difference in
reverse and forward BLEU scores;
$n =$ number of MT systems.}
\label{bleu-diff-by-lp}
\end{center}
\end{table}

As mentioned previously, however, comparison of BLEU scores is not as straightforward as human evaluation and there are further consideration to be made before drawing conclusions from the mix of positive and negative absolute BLEU score differences described above.
For example, the fact that splitting the test set into
forward and reverse directions creates two test sets
comprised of distinct sentences is likely to impact how each 
distinct BLEU score should be interpreted, as BLEU 
is not a simple arithmetic average of sentence scores 
(like human evaluation DA scores) but rather the geometric mean of 4-gram precision 
combined with a brevity penalty.
An important difference that could impact BLEU score
interpretation, for example, could be sentence length,
a large divergence resulting in forward and reverse BLEU
scores becoming not entirely comparable.

\begin{figure*}
\begin{center}
\includegraphics[width=0.85\textwidth]{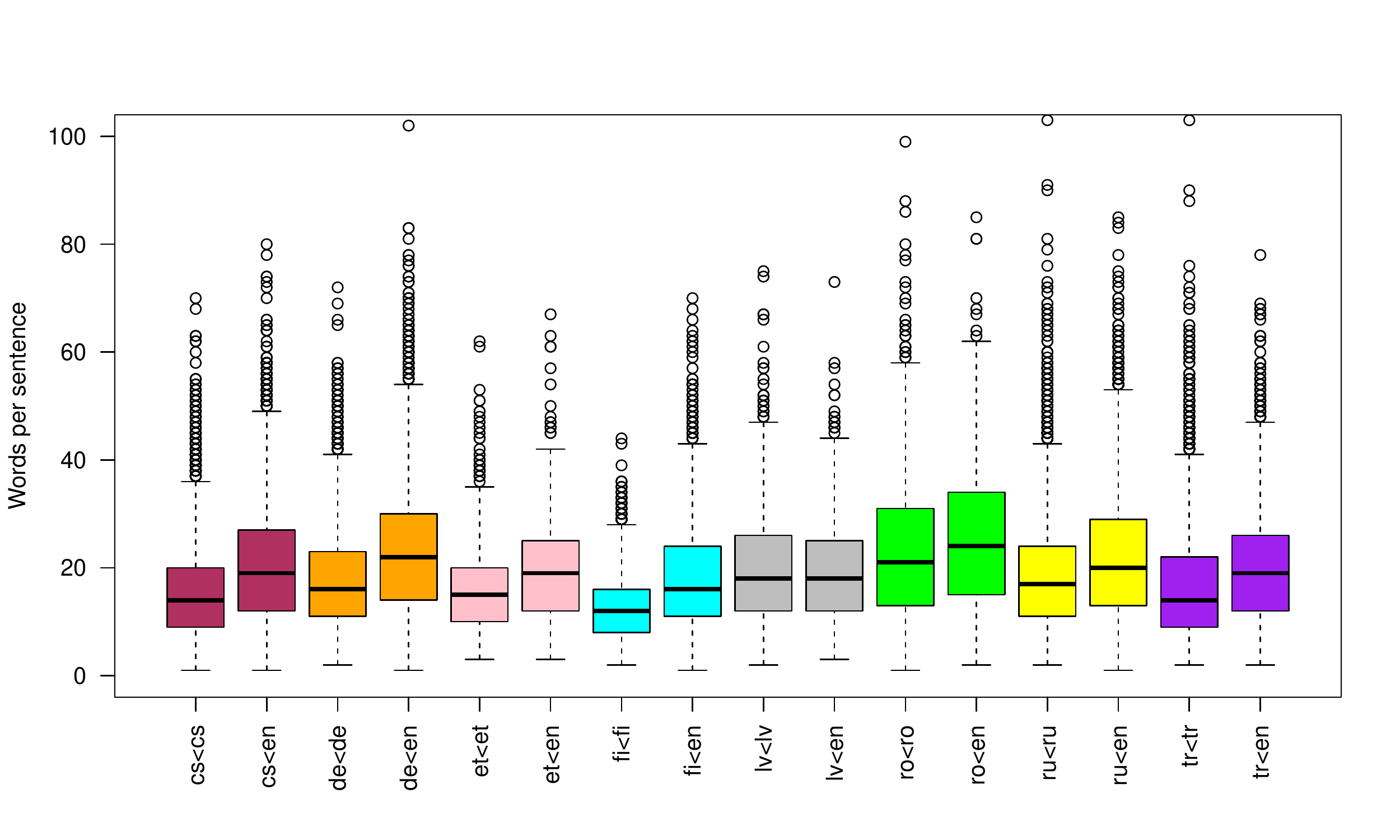}
\caption{Sentence length distribution in test data of 
{\wmt{15}}--{\wmt{18}} news translation task for text in 
non-English languages, where, for example,
cs$<$en depicts text originating in English that was manually
translated into Czech; cs$<$cs depicts text that
genuinely originated in Czech; colors depict text in
the same language; note that plots
are intentionally cropped in favor of providing better detail
of differences in median scores at the cost of omitting some
outliers. \phicomment{maybe not use "cs$<$cs" but just "cs" since here is no "$<$" happening.}}\label{words-ooen-boxplot}
\end{center}
\end{figure*}

\begin{figure}
\begin{center}
\includegraphics[width=0.5\textwidth]{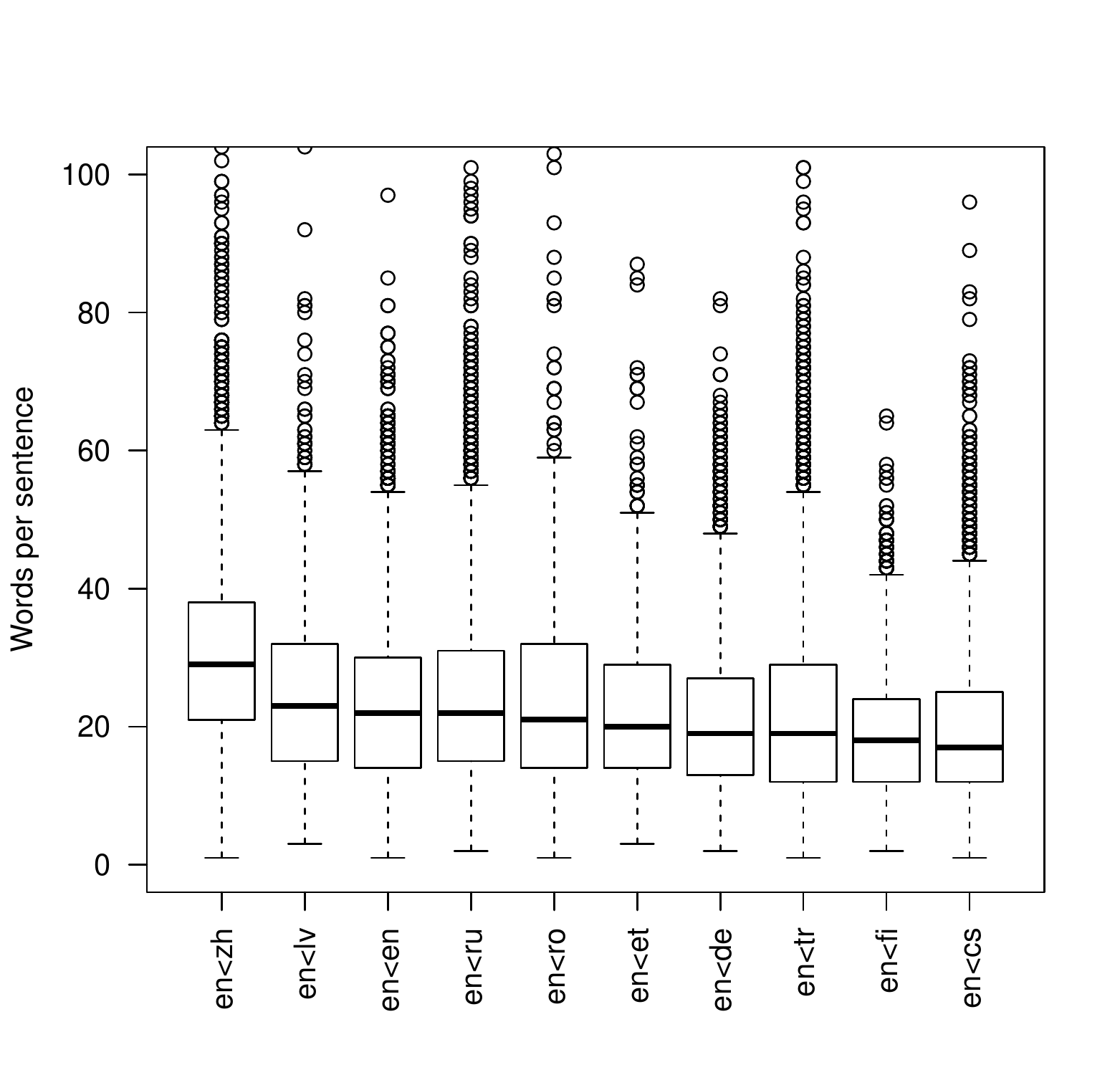}
\caption{Sentence length distribution in test data of 
{\wmt{15}}--{\wmt{18}} news translation task for text in English, 
where, for example,
en$<$cs depicts text originating in Czech that was manually
translated into English; en$<$en depicts text that
genuinely originated in English; note that plots
are intentionally cropped in favor of providing better detail
of differences in median scores at the cost of omitting some
outliers. }\label{words-toen-boxplot}
\end{center}
\end{figure}

To investigate differences in sentence length between forward and
reverse test data, Figure \ref{words-ooen-boxplot} shows 
sentence length distributions for {\wmt{15}}--{\wmt{18}} test sets firstly for all non-English languages and Figure \ref{words-toen-boxplot} shows equivalent distributions for English test sets.\footnote{Since Chinese language text has no direct equivalent to sentence length in the other languages, we omit it from this part of the analysis.}
For non-English languages (Figure \ref{words-ooen-boxplot}), there is a clear trend for text that genuinely originated in a given language to have shorter sentences than those translated from English
into that language, and this could be artifact, for example, of translated text being found to be more explicit than the  original source and less  ambiguous \cite{Bakeretal:93}.
The only exception to this trend of longer translated text exists for Latvian test data, where sentence length distributions for both text originating in Latvian and text translated from English to Latvian unusually have very similar sentence length distributions. 

For English text (Figure \ref{words-toen-boxplot}), in general sentence length distributions appear to depend on the source language,
with sentence length of text originating in English being lower than English text originating in Chinese and Latvian
but longer than text originating in all remaining non-English languages. In summary, our analysis indicates that \emph{in general translationese is shorter than text originating in a given language}.

\begin{figure*}
\begin{center}
\includegraphics[width=0.85\textwidth]{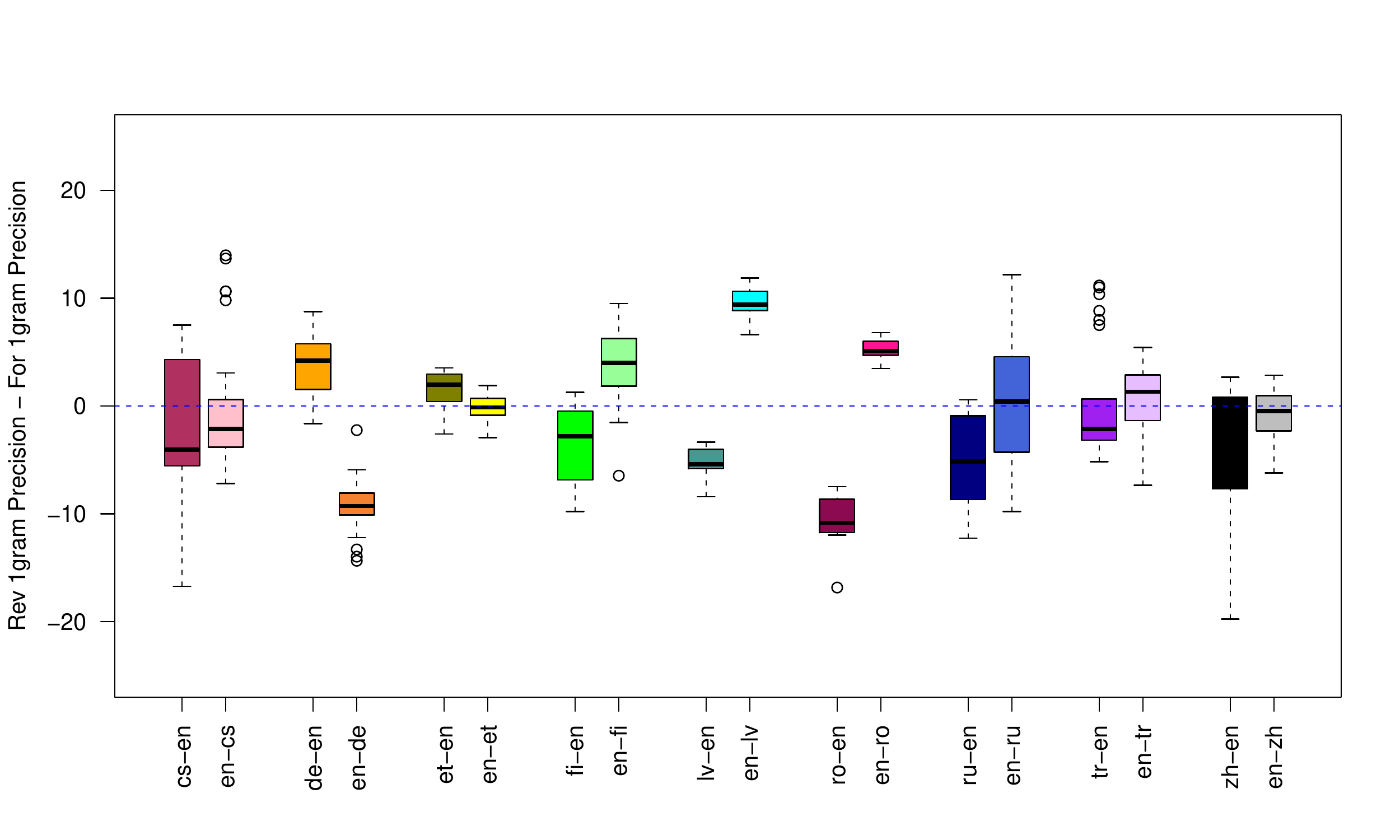}
\caption{Differences in unigram precision for systems participating
in {\wmt{15}}--{\wmt{18}} news translation task computed
for test sentences created in the reverse direction to
testing and those created in the same/forward
direction to testing broken down by language pair, again showing
a mix of positive and negative differences in scores
depending on test set creation direction, where 
``Rev 1gram Precision $-$ For 1gram Precision'' $=$
reverse unigram precision $-$ forward unigram precision.}\label{bleu-1gram}
\end{center}
\end{figure*}

Since we have observed systematic differences in sentence length
in test data for forward and reverse directions, a closer
look at interpretation of BLEU scores is necessary.
A main component of BLEU with an interpretation that does not depend on sentence length is unigram precision.
Comparison of unigram precision scores for forward and reverse test data may provide 
a better comparison therefore.
Figure \ref{bleu-1gram} shows a boxplot of unigram precision 
score differences (reverse unigram precision minus forward unigram precision) 
for {\wmt{15}}--{\wmt{18}} news task systems, and although differences in BLEU score interpretation have been removed from the comparison, there remains a clear presence of the surprisingly higher forward scores previously observed in BLEU score differences.
Our analysis indicates that the lack of observing higher reverse BLEU scores cannot be explained by sentence length and this could be an indication that our comparison of scores, even at the level of unigram precision, still has issues caused by the fact that reverse and forward test sentences are distinct. To overcome this challenge of providing a fair comparison of forward and reverse BLEU scores, in the section that follows we examine relative differences in scores for pairs of systems instead of absolute differences for individual systems. In this way, it is possible to compute BLEU score differences on the  same test sentences for pairs of systems before examining the extremity of such differences for each language direction.

\subsection{Relative Differences}

To overcome issues caused by comparison of reverse and forward BLEU being different test sentences is to compare scores primarily for reverse-created test data separately from forward-created test data.

Besides absolute differences in BLEU scores for individual systems, we should also consider how the differences in BLEU scores that occur when we change from forward to reverse test data correspond to one another, i.e. how changes in scores correspond from one system to another.
For example, for an individual competition, the problems associated with test data creation are more problematic if they occur differently for different systems participating in the same competition and less severe if they affect all systems equally, as system scores are mainly interpreted relative to one another.
To investigate this further, we examine relative changes in BLEU scores for pairs of systems, and  compare BLEU score changes for \emph{all pairs of systems} participating in the same evaluation campaign.

\begin{figure}
\begin{center}
\includegraphics[width=0.5\textwidth]{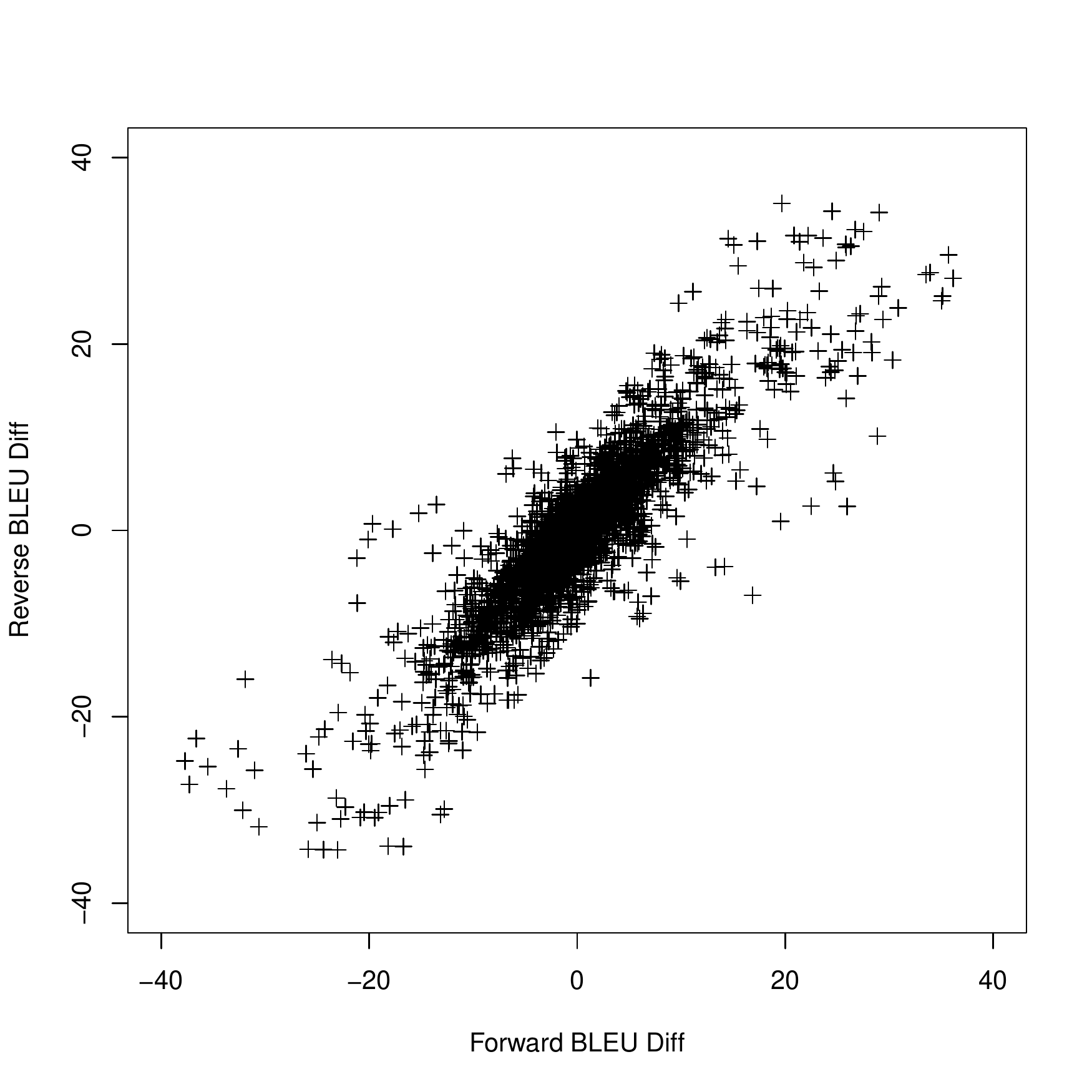}
\caption{Differences in BLEU scores for pairs of systems participating
in {\wmt{15}}--{\wmt{18}} news translation task computed
for test sentences created in the reverse and
forward directions, where 
``Reverse BLEU Diff'' $=$
reverse BLEU $-$ reverse BLEU for a pair of MT systems and ``Forward BLEU Diff'' $=$ forward BLEU $-$ forward BLEU for the same pair of MT systems.}\label{bleu-pair-diff}
\end{center}
\end{figure}

The scatter plot in Figure \ref{bleu-pair-diff} shows relative differences in BLEU scores when
we change from forward to reverse test data for all pairs of systems participating in {\wmt{15}}--{\wmt{18}}.
The absence of systems in the upper-left and lower-right quadrants reassuringly shows that although extreme changes in BLEU scores do occur when test set creation direction is altered, the changes are at least somewhat systematic in the sense that when a difference in BLEU scores occurs (a drop or increase when we change from forward to reverse test data), it occurs similarly for pairs of systems. 
However, the although there is a diagonal orientation in the plot, it still is somewhat worryingly broad and it remains possible that inclusion of reverse test data could bias BLEU scores in different ways for different types of systems.

\begin{figure}
\begin{center}
\includegraphics[width=0.5\textwidth]{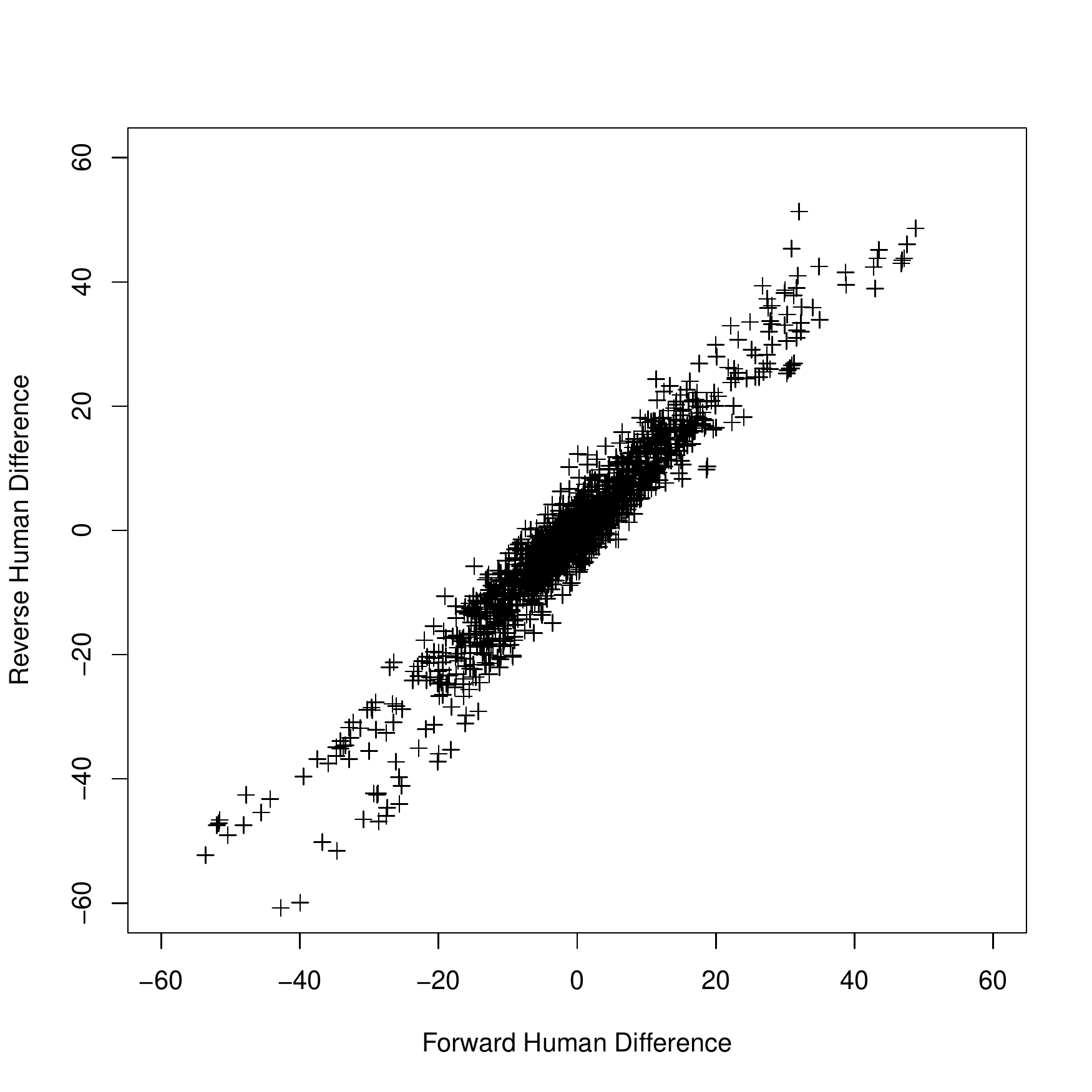}
\caption{Differences in Human evaluation DA scores for pairs of systems participating in {\wmt{17}}-{\wmt{18}} news translation task computed
for test sentences created in the reverse and forward directions, where ``Reverse Human Difference'' $=$
reverse DA $-$ reverse DA for a pair of MT systems and ``Forward Human Difference'' $=$ forward DA $-$ forward DA for the same pair of MT systems.}\label{human-pair-diff}
\end{center}
\end{figure}

In terms of human evaluation, the scatter plot in Figure \ref{human-pair-diff} shows relative  differences in human scores for all pairs of systems in {\wmt{17}}--{\wmt{18}}. Again, the absence of systems in the upper-left and lower-right
quadrants shows a similar trend for human evaluation, where relative differences in DA scores for pairs of systems correspond very closely when we change from reverse to forward-created test data.

In summary, examination of absolute differences in human scores revealed almost across the board higher human scores when systems are tested on data created in the reverse direction to testing,
while BLEU scores showed a mix of higher and lower reverse scores for test set creation directions.
Although it is important to understand the changes in absolute scores that should be expected, relative differences in performance are more important, as these directly impact conclusions about what systems and methods outperform others. More reassuring than absolute differences in BLEU scores, relative differences correspond quite well between pairs of competing systems. 
The correspondence of relative differences for pairs of systems was extremely close for human evaluation and this provides evidence of the validity of conclusions made in past human evaluations of MT that included reverse test data. 
However the spread in \ref{bleu-pair-diff} appear to suggest 
that system rankings could still change if we measure forward vs reverse BLEU, and although the corresponding DA graph (Figure \ref{human-pair-diff} is much narrower (which is good) but there still could be changes in ranking. We therefore include further analysis that provides a direct comparison of system rankings for past evaluations in terms of both BLEU and human evaluation.

\section{System Rankings}

Tables \ref{tab:bleu_sys_rank}(a), \ref{tab:bleu_sys_rank}(b) and \ref{tab:bleu_sys_rank}(c) show the Kendall's $\tau$ rank correlation of forward and reverse BLEU scores for systems participating in  \wmt{15}--\wmt{18} individual competitions, in addition to Pearson and Spearman correlations of same. 
As can be seen, the correspondence between forward and reverse rank correlation of systems according to BLEU varies considerably across different evaluation test sets, from as low as a $\tau$ of $0.2$, where BLEU score rankings are extremely different depending on test data creation direction, up to a $\tau$ of $1.0$, where rank correlation is identical (cs-en; fi-en newstest2017; fi-en; en-cs newstest2018).

\begin{table}
    \centering
    \input{tbl-02-bleu-sys-ranking.tex}
    \caption{Pearson ($r$), Spearman ($\rho$) and Kendall's $\tau$ correlation of forward and reverse BLEU scores of all systems participating in {\wmt{15}} -- {\wmt{18}} news translation task; language pairs ordered from lowest to highest Pearson correlation.}
    \label{tab:bleu_sys_rank}
\end{table}

Similarly, Table \ref{tab:human_sys_rank} shows the correlation of rankings of human evaluation data, where Kendall's $\tau$ correlations of forward and reverse test data also range from little correspondence for tr-en newstest2018 at 0.4 in the worst case to identical system rankings $\tau$ of $1.0$in five cases (cs-en; fin-en; en-tr newstest2017; en-ru; en-cs newstest2018).

\begin{table}
    \centering
    \input{tbl-03-human-sys-ranking.tex}
    \caption{Pearson ($r$), Spearman ($\rho$) and Kendall's $\tau$ correlation of forward and reverse Human DA scores of all systems participating in {\wmt{17}} -- {\wmt{18}} news translation task; language pairs ordered from lowest to highest Pearson correlation.}
    \label{tab:human_sys_rank}
\end{table}

In overall summary, our analysis of differences in both BLEU and human evaluation scores reveal differences in system rankings when tested on reverse and forward-created test data, differences substantial in some cases. Subsequently we have confirmed the validity of suspicions raised about potential  lack of reliability of test data raised by \newcite{Toraletal:18} and \newcite{laeublietal:18} caused by inclusion of reverse-created test data. However, as stated previously, both reassessments of \newcite{hassanetal:18} include the

%% file: tbl-01-humandiff-by-lp.tex
\footnotesize
\begin{tabular}{crrrrrr}
\toprule
& R$>$F   &  F$>$R  & R$-$F                     & R$-$F 						& \multicolumn{1}{c}{$n$} \\
& (\%)    & (\%)    & \multicolumn{1}{c}{$\mu$} & \multicolumn{1}{c}{$\sigma$} 	&                         \\

\midrule      

\multicolumn{1}{r}{et-en} &  92.9 & 7.1 & 7.03 & 3.26 & 14 \\
\multicolumn{1}{r}{zh-en} &  93.3 & 6.7 & 3.56 & 2.21 & 30 \\
\multicolumn{1}{r}{ru-en} &  94.1 & 5.9 & 1.65 & 2.08 & 17 \\
\multicolumn{1}{r}{cs-en} & 100.0 & 0.0 & 5.21 & 3.00 &  9 \\
\multicolumn{1}{r}{de-en} & 100.0 & 0.0 & 7.73 & 2.52 & 27 \\
\multicolumn{1}{r}{fi-en} & 100.0 & 0.0 & 4.48 & 2.29 & 15 \\
\multicolumn{1}{r}{lv-en} & 100.0 & 0.0 & 3.17 & 1.89 &  9 \\
\multicolumn{1}{r}{tr-en} & 100.0 & 0.0 & 5.25 & 2.23 & 15 \\[1ex]

\multicolumn{1}{l}{en-fi} &  95.8 & 4.2 &  9.47 & 4.62 & 24 \\
\multicolumn{1}{l}{en-de} &  96.9 & 3.1 &  6.53 & 3.40 & 32 \\
\multicolumn{1}{l}{en-cs} & 100.0 & 0.0 & 12.67 & 2.82 & 20 \\
\multicolumn{1}{l}{en-et} & 100.0 & 0.0 &  6.21 & 1.76 & 14 \\
\multicolumn{1}{l}{en-lv} & 100.0 & 0.0 & 18.04 & 4.10 & 17 \\
\multicolumn{1}{l}{en-ru} & 100.0 & 0.0 & 13.28 & 3.08 & 18 \\
\multicolumn{1}{l}{en-tr} & 100.0 & 0.0 & 11.38 & 7.24 & 16 \\
\multicolumn{1}{l}{en-zh} & 100.0 & 0.0 &  8.43 & 2.06 & 25 \\

\bottomrule
\end{tabular}

%% file: tbl-00-bleudiff-by-lp.tex
\footnotesize
\begin{tabular}{crrrrrr}
\toprule
& R$>$F   &  F$>$R  & R$-$F                     & R$-$F 						& \multicolumn{1}{c}{$n$} \\
& (\%)    & (\%)    & \multicolumn{1}{c}{$\mu$} & \multicolumn{1}{c}{$\sigma$} 	&                         \\

\midrule      

\multicolumn{1}{r}{lv-en} &  11.1 & 88.9 & $-$2.00 & 2.12 &  9 \\
\multicolumn{1}{r}{zh-en} &  33.3 & 66.7 & $-$2.52 & 6.54 & 30 \\
\multicolumn{1}{r}{fi-en} &  42.1 & 57.9 & $-$0.06 & 2.85 & 38 \\
\multicolumn{1}{r}{ru-en} &  47.5 & 52.5 &    0.90 & 5.53 & 40 \\
\multicolumn{1}{r}{cs-en} &  48.6 & 51.4 &    1.13 & 6.78 & 37 \\
\multicolumn{1}{r}{tr-en} &  76.0 & 24.0 &    4.38 & 5.45 & 25 \\
\multicolumn{1}{r}{et-en} &  78.6 & 21.4 &    2.30 & 2.12 & 14 \\
\multicolumn{1}{r}{de-en} & 100.0 &  0.0 &   10.03 & 4.92 & 50 \\[1ex]

\multicolumn{1}{l}{en-de} &   1.6 & 98.4 & $-$6.34 & 3.39 & 63 \\
\multicolumn{1}{l}{en-zh} &  36.0 & 60.0 &    0.02 & 1.63 & 25 \\
\multicolumn{1}{l}{en-cs} &  52.7 & 47.3 &    0.56 & 3.35 & 55 \\
\multicolumn{1}{l}{en-ru} &  65.0 & 35.0 &    3.09 & 5.82 & 40 \\
\multicolumn{1}{l}{en-et} &  71.4 & 28.6 &    0.86 & 1.21 & 14 \\
\multicolumn{1}{l}{en-tr} &  84.0 & 12.0 &    2.53 & 2.56 & 25 \\
\multicolumn{1}{l}{en-fi} &  87.2 & 12.8 &    3.07 & 2.31 & 47 \\
\multicolumn{1}{l}{en-lv} & 100.0 &  0.0 &    8.12 & 2.50 & 17 \\

\bottomrule
\end{tabular}

%% file: tbl-02-bleu-sys-ranking.tex
\scriptsize

\begin{tabular}{lccc}
\toprule
newstest15      & $r$   & $\rho$& $\tau$ \\
\midrule
en-ru & 0.838 & 0.498 & 0.405 \\ 
fi-en & 0.900 & 0.873 & 0.670 \\ 
ru-en & 0.903 & 0.934 & 0.821 \\ 
en-fi & 0.911 & 0.842 & 0.689 \\ 
en-de & 0.932 & 0.891 & 0.717 \\ 
de-en & 0.952 & 0.879 & 0.769 \\ 
cs-en & 0.985 & 0.832 & 0.717 \\ 
en-cs & 0.995 & 0.963 & 0.880 \\[2ex]
\midrule
newstest16      & $r$   & $\rho$& $\tau$ \\
\midrule
ro-en & 0.489 & 0.679 & 0.524 \\ 
en-de & 0.787 & 0.545 & 0.421 \\ 
tr-en & 0.795 & 0.783 & 0.667 \\ 
en-ru & 0.809 & 0.252 & 0.182 \\ 
fi-en & 0.845 & 0.820 & 0.648 \\ 
en-fi & 0.875 & 0.889 & 0.735 \\ 
en-tr & 0.881 & 0.850 & 0.722 \\ 
en-ro & 0.945 & 0.930 & 0.818 \\ 
ru-en & 0.945 & 0.697 & 0.600 \\ 
en-cs & 0.954 & 0.598 & 0.466 \\ 
de-en & 0.958 & 0.818 & 0.644 \\ 
cs-en & 0.961 & 0.655 & 0.504 \\
\midrule
newstest17      & $r$   & $\rho$& $\tau$ \\
\midrule
en-zh & 0.608 & 0.601 & 0.367 \\ 
zh-en & 0.646 & 0.838 & 0.667 \\ 
en-lv & 0.861 & 0.860 & 0.735 \\ 
cs-en & 0.865 & 1.000 & 1.000 \\ 
lv-en & 0.879 & 0.883 & 0.778 \\ 
en-ru & 0.890 & 0.750 & 0.667 \\ 
tr-en & 0.901 & 0.927 & 0.778 \\ 
en-de & 0.933 & 0.718 & 0.567 \\ 
de-en & 0.937 & 0.836 & 0.673 \\ 
en-tr & 0.939 & 0.976 & 0.929 \\ 
ru-en & 0.942 & 0.817 & 0.611 \\ 
en-cs & 0.961 & 0.945 & 0.842 \\ 
en-fi & 0.969 & 0.944 & 0.848 \\ 
fi-en & 0.988 & 1.000 & 1.000 \\
\midrule
newstest18      & $r$   & $\rho$& $\tau$ \\
\midrule
tr-en & 0.106 & 0.314 & 0.200 \\ 
en-zh & 0.570 & 0.445 & 0.333 \\ 
cs-en & 0.579 & 0.700 & 0.600 \\ 
zh-en & 0.771 & 0.616 & 0.442 \\ 
en-tr & 0.897 & 0.611 & 0.546 \\ 
en-de & 0.938 & 0.741 & 0.583 \\ 
de-en & 0.954 & 0.897 & 0.750 \\ 
fi-en & 0.963 & 1.000 & 1.000 \\ 
en-et & 0.966 & 0.978 & 0.912 \\ 
en-ru & 0.966 & 0.983 & 0.944 \\ 
ru-en & 0.966 & 0.857 & 0.714 \\ 
en-fi & 0.981 & 0.986 & 0.939 \\ 
et-en & 0.985 & 0.978 & 0.912 \\ 
en-cs & 0.990 & 1.000 & 1.000 \\ 

\bottomrule
\end{tabular}

%% file: tbl-03-human-sys-ranking.tex
\scriptsize

\begin{tabular}{lccc}
\toprule
newstest17      & $r$   & $\rho$& $\tau$ \\
\midrule
zh-en & 0.935 & 0.903 & 0.758 \\ 
ru-en & 0.939 & 0.883 & 0.778 \\ 
de-en & 0.949 & 0.909 & 0.782 \\ 
en-cs & 0.952 & 0.952 & 0.857 \\ 
en-lv & 0.952 & 0.904 & 0.765 \\ 
cs-en & 0.957 & 1.000 & 1.000 \\ 
en-ru & 0.958 & 0.817 & 0.722 \\ 
lv-en & 0.972 & 0.967 & 0.889 \\ 
en-zh & 0.977 & 0.939 & 0.822 \\ 
en-de & 0.979 & 0.921 & 0.771 \\ 
fi-en & 0.979 & 1.000 & 1.000 \\ 
tr-en & 0.983 & 0.927 & 0.822 \\ 
en-fi & 0.989 & 0.902 & 0.788 \\ 
en-tr & 0.992 & 1.000 & 1.000 \\
\midrule
newstest18      & $r$   & $\rho$& $\tau$ \\
\midrule
tr-en & 0.703 & 0.600 & 0.400 \\ 
en-tr & 0.865 & 0.619 & 0.500 \\ 
zh-en & 0.884 & 0.644 & 0.495 \\ 
fi-en & 0.898 & 0.667 & 0.556 \\ 
en-fi & 0.959 & 0.902 & 0.758 \\ 
en-ru & 0.969 & 1.000 & 1.000 \\ 
en-cs & 0.974 & 1.000 & 1.000 \\ 
et-en & 0.975 & 0.925 & 0.846 \\ 
cs-en & 0.978 & 0.900 & 0.800 \\ 
en-zh & 0.981 & 0.969 & 0.890 \\ 
ru-en & 0.983 & 0.976 & 0.929 \\ 
de-en & 0.984 & 0.865 & 0.750 \\ 
en-et & 0.984 & 0.974 & 0.890 \\ 
en-de & 0.990 & 0.947 & 0.850 \\ 

\bottomrule
\end{tabular}

%% file: sec-05-reeval.tex
\section{Re-evaluation of Human Parity Claims}

As mentioned previously in Section \ref{relwork}, past re-evaluations of human parity claims were hampered by low inter-annotator agreement levels, employment of older human evaluation technologies than the original, treatment of Trueskill clusters to draw conclusions of statistical significance and lack of statistical power analysis for planned sample size, while the original evaluation itself suffered severely from inclusion of reverse-created data we have shown to be problematic, as well as a very low number of distinct translations included in the evaluation.

In our re-evaluation of the original, we firstly carry out statistical power analysis so that in the case of encountering any ties between systems or indeed human and system, that tests used to draw conclusions have sufficient statistical power to avoid human-parity claims that in fact simply correspond to a Type II error. Statistical power is of particular importance when considering document-level evaluation due to the fact that gathering ratings of documents as opposed to sentences requires substantially more annotator time and for this reason is likely to result in a reduction in the number of assessments collected in any evaluation. 
For example, \newcite{laeublietal:18} included as few as 55 documents in their re-evaluation of \newcite{hassanetal:18}. Our concern about a potential substantial reduction in sample size in future document-level evaluations is well-founded therefore, especially considering standard segment-level MT human evaluations commonly include a sample of 1,500 segments. In the case of \newcite{laeublietal:18} this corresponds to an extreme reduction of approximately  96\% to the sample size. Since the very nature of the question being investigated involves a potential tie between human and machine, such a small sample size is a serious risk to the reliability of conclusions drawn simply due to its impact in terms of statistical power. 

For this reason, prior to running our re-evaluation, we run power analysis to investigate an appropriate sample size that will result in sufficiently powerful tests. 
As a rough guide to what constitutes sufficient statistical power, we borrow the five-eighty convention from the  behavioural sciences that provides a balance between Type I versus Type II error, where significance and power levels are set at 0.05 and 0.8 respectively \cite{cohen88}.

Table \ref{power} shows the statistical power, the probability of identifying a significant difference when one exists,
of the statistical test applied in {\wsmt}
evaluations, Wilcoxon rank-sum test, for
a range of effect and sample sizes ($n$),
where for the purpose of the test the appropriate effect size is the probability of the translations of system A being scored lower than those
of system B. As shown in Table \ref{power}
for the usual sample size employed in {\wsmt} evaluations, 1,500, statistical power even for closely performing systems, where the probability of the translations of system A being scored  lower  than  those  of  system  B is 0.47, statistical power is still above 0.8.
For such pairs of systems, however, if we were to employ the smaller sample size of 55 documents, as in \newcite{laeublietal:18}, the power of the test to identify a significant differences falls as low as 0.081, approaching one tenth of acceptable statistical power levels.\footnote{In \newcite{laeublietal:18} the Sign test was used as opposed to Wilcoxon rank sum and has similar statistical power for such an effect size.}

\begin{table*}[]
    \centering
    \input{tbl-05-power}
    \caption{Statistical Power of two-sided Wilcoxon Rank Sum Test for a range of sample and effect sizes; power $\geq$ 0.8 highlighted in bold.}
    \label{power}
\end{table*}

In order to further put into context the closeness in human performance of systems we can expect to encounter in our planned re-evaluation, we examine the effect size for pairs of systems in the original. Table \ref{hassaneffectsize} shows the effect size for all pairs of systems included in \newcite{hassanetal:18}. If we take, for example, the effect size between the top two runs, Ref-HT and Combo-5 of 0.435, we can roughly see from Table \ref{power} that the likelihood of identifying a significant difference at this effect size ranges from as low as 0.188 for a sample size of 55 and only reaches an acceptable level above 0.8 at sample size 385. Since the test set used in \newcite{hassanetal:18} included a far lower number of test documents however, basing our evaluation on document ratings would lead to low statistical power and likely result in Type II errors cause by this small sample size. 

A good compromise between fully document-level evaluation, where only ratings of documents are collected, and fully segment-level  evaluations, in which segments are presented to human judges in isolation of the document, is collection of ratings of segments with the wider document context available to the human assessor and have the segments evaluated in their original order. In this way, a sufficient sample size can still be achieved to ensure appropriate levels of statistical power with the added aim of human judges being able to take into account the quality of translations within the wider document context. Although \newcite{Toraletal:18} did not specifically indicate statistical power analysis as their particular motivation, this segment-rating document-context approach appears to be that which they employed.

\begin{table*}[]
    \centering
    \input{tbl-07-hassan-effectsize}
    \caption{Effect size, probability of a translation produced by the system in a given row receiving a lower DA score than that of the system in a given column; systems and data taken from \newcite{hassanetal:18} human evaluation.}
    \label{hassaneffectsize}
\end{table*}

We therefore plan our re-evaluation as follows:
\begin{itemize}
    \item Collect segment ratings for documents produced by a single system within the correct document context;
    \item Aim to collect direct assessments of a sufficient number of translations exceeding the minimum acceptable sample size in terms of power analysis, approximately 385 distinct translations; 
    \item Use $n$, the number of distinct translations as opposed to repeat human assessments as the sample size;
    \item Employ Direct Assessment, the most up to date technology for this purpose and that employed by {\wsmt} for the official results since 2017, a method shown to produce highly repeatable results;
    \item Only employ forward-created test data;
    \item Only draw conclusions specific to Chinese to English translation and news domain;
    \item Produce clusters with a standard significance test, Wilcoxon rank-sum test.
\end{itemize}

\subsection{Re-evaluation Results}

\begin{table*}[t]
    \centering
    \input{tbl-06-hassan-reeval}
    \caption{Re-evaluation of human-parity-claimed Chinese to English system of \newcite{hassanetal:18}; {\onesig} denotes system that significantly outperforms all lower ranked systems according to a two-sided Wilcoxon rank-sum test  $p<0.05$}
    \label{hassanreeval}
\end{table*}

Direct Assessment (DA) HITs were set up and run as in {\wsmt}
human evaluations on Mechanical Turk but with the distinction of segments being evaluated in the correct order in which they
appeared in a document, comprising an initial set of results, which we refer to as segment rating $+$ document context (SR$+$DC). In addition to the segment rating workers were additionally shown entire documents and asked to rate them, providing a secondary set of results for comparison purposes. We refer to these fully document-level results as document rating $+$ document context (DR$+$DC) configuration. As is usual in DA evaluations, translations were rated in a 0--100 rating scale and quality control was applied. 

131 workers participated producing a total of 13,214 assessments of translations, of which 6,606 (49.99\%) were from workers who passed DA's quality control checks.

Table \ref{hassanreeval} shows results of our re-evaluation of the top systems originally included in \newcite{hassanetal:18}, where REF-HT is the original set of human translations produced by \newcite{hassanetal:18} themselves and  against which human-parity of MT was claimed, while REF-PE is machine translated outputs that have been post-edited by humans, and Combo-6 is the best-performing system in  \newcite{hassanetal:18}.

Results when segments are rated by human judges within the 
correct document context (Segment Rating + Document Context) 
show that the DA score achieved by the human reference 
translation, REF-HT, is significantly higher than both REF-PE and Combo-6, agreeing with results of both \newcite{laeublietal:18}
and 
\newcite{Toraletal:18}. 
Since this approach has a large 
enough 
sample size to ensure sufficient statistical power, the tie 
between REF-PE and Combo-6 is a legitimate one however. 
Although this tie does indeed indicate high performance of Combo-6, since REF-PE is in fact post-edited MT output however, this tie does not provide legitimate evidence to support a human-parity claim. 

Although we already know from the power analysis carried out prior to planning the evaluation that fully document-level evaluations that ask human assessors to rate documents as opposed to segments will encounter problems when ties occur, we nonetheless run this kind of evaluation for demonstration purposes. Document Rating $+$ Document Context results in Table \ref{hassanreeval} do indeed produce what appears to be a statistical tie between the three sets of outputs as none significantly outperforms all lower ranking systems. However, a conclusion of human parity cannot legitimately be claimed from this tie due to the low statistical power of the test due to the small sample of documents that were rated. Ties in this case do not indicate human-parity but simply that the test is too weak to identify significant differences between systems.

In summary, similar to \newcite{Toraletal:18} and \newcite{laeublietal:18}, our results show evidence that the original system, Combo-6, was outperformed by human translation.
It should be noted however that from our results it cannot be inferred that machine translation in general has not yet reached human performance but simply that the system that originally claimed human-parity in fact did not achieve it, as tested on data from WMT 2017 news task.

%% file: tbl-05-power.tex
\tiny
\begin{tabular}{rlllllllllllllllllllllllllllllllllllllllllllllllllllllllllllllllllllllllllllllllllllllllllllllllllllllllllllllllllllllllllllllllllllllllllllll}
\toprule 
     & & \multicolumn{17}{c}{effect size} \\

&  & 0.330 & 0.340 & 0.350 & 0.360 & 0.370 & 0.380 & 0.390 & 0.400 & 0.410 & 0.420 & 0.430 & 0.440 & 0.450 & 0.460 & 0.470 & 0.480 & 0.490 \\ 

\multicolumn{1}{c}{$n$}  &  \\
\midrule

55 &  & \bf{0.886} & \bf{0.842} & 0.788 & 0.725 & 0.659 & 0.586 & 0.512 & 0.438 & 0.367 & 0.300 & 0.243 & 0.188 & 0.144 & 0.111 & 0.081 & 0.066 & 0.056 \\ 
110 &  & \bf{0.995} & \bf{0.989} & \bf{0.977} & \bf{0.957} & \bf{0.923} & \bf{0.877} & \bf{0.813} & 0.730 & 0.639 & 0.537 & 0.434 & 0.336 & 0.246 & 0.176 & 0.120 & 0.080 & 0.057 \\ 
165 &  & \bf{1.000} & \bf{1.000} & \bf{0.998} & \bf{0.995} & \bf{0.986} & \bf{0.969} & \bf{0.941} & \bf{0.887} & \bf{0.812} & 0.714 & 0.595 & 0.470 & 0.348 & 0.242 & 0.156 & 0.095 & 0.060 \\ 
220 &  & \bf{1.000} & \bf{1.000} & \bf{1.000} & \bf{0.999} & \bf{0.998} & \bf{0.994} & \bf{0.981} & \bf{0.957} & \bf{0.911} & \bf{0.830} & 0.723 & 0.586 & 0.442 & 0.307 & 0.192 & 0.111 & 0.063 \\ 
275 &  & \bf{1.000} & \bf{1.000} & \bf{1.000} & \bf{1.000} & \bf{1.000} & \bf{0.999} & \bf{0.995} & \bf{0.984} & \bf{0.959} & \bf{0.903} & \bf{0.810} & 0.684 & 0.528 & 0.367 & 0.230 & 0.128 & 0.070 \\ 
330 &  & \bf{1.000} & \bf{1.000} & \bf{1.000} & \bf{1.000} & \bf{1.000} & \bf{1.000} & \bf{0.999} & \bf{0.995} & \bf{0.982} & \bf{0.947} & \bf{0.878} & 0.763 & 0.604 & 0.427 & 0.265 & 0.144 & 0.073 \\ 
385 &  & \bf{1.000} & \bf{1.000} & \bf{1.000} & \bf{1.000} & \bf{1.000} & \bf{1.000} & \bf{1.000} & \bf{0.998} & \bf{0.992} & \bf{0.971} & \bf{0.924} & \bf{0.824} & 0.672 & 0.485 & 0.302 & 0.159 & 0.077 \\ 
440 &  & \bf{1.000} & \bf{1.000} & \bf{1.000} & \bf{1.000} & \bf{1.000} & \bf{1.000} & \bf{1.000} & \bf{0.999} & \bf{0.997} & \bf{0.986} & \bf{0.951} & \bf{0.870} & 0.730 & 0.538 & 0.338 & 0.176 & 0.081 \\ 
495 &  & \bf{1.000} & \bf{1.000} & \bf{1.000} & \bf{1.000} & \bf{1.000} & \bf{1.000} & \bf{1.000} & \bf{1.000} & \bf{0.999} & \bf{0.992} & \bf{0.970} & \bf{0.906} & 0.778 & 0.587 & 0.372 & 0.190 & 0.087 \\ 
550 &  & \bf{1.000} & \bf{1.000} & \bf{1.000} & \bf{1.000} & \bf{1.000} & \bf{1.000} & \bf{1.000} & \bf{1.000} & \bf{0.999} & \bf{0.996} & \bf{0.982} & \bf{0.933} & \bf{0.822} & 0.632 & 0.406 & 0.210 & 0.090 \\ 
605 &  & \bf{1.000} & \bf{1.000} & \bf{1.000} & \bf{1.000} & \bf{1.000} & \bf{1.000} & \bf{1.000} & \bf{1.000} & \bf{1.000} & \bf{0.998} & \bf{0.989} & \bf{0.952} & \bf{0.855} & 0.675 & 0.439 & 0.225 & 0.093 \\ 
660 &  & \bf{1.000} & \bf{1.000} & \bf{1.000} & \bf{1.000} & \bf{1.000} & \bf{1.000} & \bf{1.000} & \bf{1.000} & \bf{1.000} & \bf{0.999} & \bf{0.993} & \bf{0.966} & \bf{0.882} & 0.713 & 0.471 & 0.241 & 0.093 \\ 
715 &  & \bf{1.000} & \bf{1.000} & \bf{1.000} & \bf{1.000} & \bf{1.000} & \bf{1.000} & \bf{1.000} & \bf{1.000} & \bf{1.000} & \bf{1.000} & \bf{0.996} & \bf{0.977} & \bf{0.908} & 0.745 & 0.502 & 0.257 & 0.101 \\ 
770 &  & \bf{1.000} & \bf{1.000} & \bf{1.000} & \bf{1.000} & \bf{1.000} & \bf{1.000} & \bf{1.000} & \bf{1.000} & \bf{1.000} & \bf{1.000} & \bf{0.998} & \bf{0.983} & \bf{0.926} & 0.775 & 0.531 & 0.273 & 0.105 \\ 
825 &  & \bf{1.000} & \bf{1.000} & \bf{1.000} & \bf{1.000} & \bf{1.000} & \bf{1.000} & \bf{1.000} & \bf{1.000} & \bf{1.000} & \bf{1.000} & \bf{0.999} & \bf{0.989} & \bf{0.942} & \bf{0.804} & 0.560 & 0.288 & 0.107 \\ 
880 &  & \bf{1.000} & \bf{1.000} & \bf{1.000} & \bf{1.000} & \bf{1.000} & \bf{1.000} & \bf{1.000} & \bf{1.000} & \bf{1.000} & \bf{1.000} & \bf{0.999} & \bf{0.992} & \bf{0.954} & \bf{0.829} & 0.587 & 0.307 & 0.108 \\ 
935 &  & \bf{1.000} & \bf{1.000} & \bf{1.000} & \bf{1.000} & \bf{1.000} & \bf{1.000} & \bf{1.000} & \bf{1.000} & \bf{1.000} & \bf{1.000} & \bf{1.000} & \bf{0.995} & \bf{0.964} & \bf{0.848} & 0.613 & 0.321 & 0.118 \\ 
990 &  & \bf{1.000} & \bf{1.000} & \bf{1.000} & \bf{1.000} & \bf{1.000} & \bf{1.000} & \bf{1.000} & \bf{1.000} & \bf{1.000} & \bf{1.000} & \bf{1.000} & \bf{0.997} & \bf{0.971} & \bf{0.868} & 0.638 & 0.338 & 0.121 \\ 
1045 &  & \bf{1.000} & \bf{1.000} & \bf{1.000} & \bf{1.000} & \bf{1.000} & \bf{1.000} & \bf{1.000} & \bf{1.000} & \bf{1.000} & \bf{1.000} & \bf{1.000} & \bf{0.998} & \bf{0.977} & \bf{0.886} & 0.661 & 0.354 & 0.123 \\ 
1100 &  & \bf{1.000} & \bf{1.000} & \bf{1.000} & \bf{1.000} & \bf{1.000} & \bf{1.000} & \bf{1.000} & \bf{1.000} & \bf{1.000} & \bf{1.000} & \bf{1.000} & \bf{0.998} & \bf{0.983} & \bf{0.902} & 0.684 & 0.370 & 0.127 \\ 
1155 &  & \bf{1.000} & \bf{1.000} & \bf{1.000} & \bf{1.000} & \bf{1.000} & \bf{1.000} & \bf{1.000} & \bf{1.000} & \bf{1.000} & \bf{1.000} & \bf{1.000} & \bf{0.999} & \bf{0.987} & \bf{0.916} & 0.703 & 0.384 & 0.131 \\ 
1210 &  & \bf{1.000} & \bf{1.000} & \bf{1.000} & \bf{1.000} & \bf{1.000} & \bf{1.000} & \bf{1.000} & \bf{1.000} & \bf{1.000} & \bf{1.000} & \bf{1.000} & \bf{0.999} & \bf{0.990} & \bf{0.928} & 0.725 & 0.398 & 0.133 \\ 
1265 &  & \bf{1.000} & \bf{1.000} & \bf{1.000} & \bf{1.000} & \bf{1.000} & \bf{1.000} & \bf{1.000} & \bf{1.000} & \bf{1.000} & \bf{1.000} & \bf{1.000} & \bf{1.000} & \bf{0.992} & \bf{0.936} & 0.745 & 0.413 & 0.138 \\ 
1320 &  & \bf{1.000} & \bf{1.000} & \bf{1.000} & \bf{1.000} & \bf{1.000} & \bf{1.000} & \bf{1.000} & \bf{1.000} & \bf{1.000} & \bf{1.000} & \bf{1.000} & \bf{1.000} & \bf{0.994} & \bf{0.947} & 0.760 & 0.429 & 0.145 \\ 
1375 &  & \bf{1.000} & \bf{1.000} & \bf{1.000} & \bf{1.000} & \bf{1.000} & \bf{1.000} & \bf{1.000} & \bf{1.000} & \bf{1.000} & \bf{1.000} & \bf{1.000} & \bf{1.000} & \bf{0.995} & \bf{0.954} & 0.777 & 0.443 & 0.148 \\ 
1430 &  & \bf{1.000} & \bf{1.000} & \bf{1.000} & \bf{1.000} & \bf{1.000} & \bf{1.000} & \bf{1.000} & \bf{1.000} & \bf{1.000} & \bf{1.000} & \bf{1.000} & \bf{1.000} & \bf{0.997} & \bf{0.960} & 0.794 & 0.457 & 0.155 \\ 
1485 &  & \bf{1.000} & \bf{1.000} & \bf{1.000} & \bf{1.000} & \bf{1.000} & \bf{1.000} & \bf{1.000} & \bf{1.000} & \bf{1.000} & \bf{1.000} & \bf{1.000} & \bf{1.000} & \bf{0.997} & \bf{0.965} & \bf{0.809} & 0.471 & 0.156 \\ 
1540 &  & \bf{1.000} & \bf{1.000} & \bf{1.000} & \bf{1.000} & \bf{1.000} & \bf{1.000} & \bf{1.000} & \bf{1.000} & \bf{1.000} & \bf{1.000} & \bf{1.000} & \bf{1.000} & \bf{0.998} & \bf{0.971} & \bf{0.821} & 0.485 & 0.161 \\ 
1595 &  & \bf{1.000} & \bf{1.000} & \bf{1.000} & \bf{1.000} & \bf{1.000} & \bf{1.000} & \bf{1.000} & \bf{1.000} & \bf{1.000} & \bf{1.000} & \bf{1.000} & \bf{1.000} & \bf{0.998} & \bf{0.975} & \bf{0.838} & 0.499 & 0.164 \\ 
1650 &  & \bf{1.000} & \bf{1.000} & \bf{1.000} & \bf{1.000} & \bf{1.000} & \bf{1.000} & \bf{1.000} & \bf{1.000} & \bf{1.000} & \bf{1.000} & \bf{1.000} & \bf{1.000} & \bf{0.999} & \bf{0.979} & \bf{0.846} & 0.512 & 0.171 \\ 
1705 &  & \bf{1.000} & \bf{1.000} & \bf{1.000} & \bf{1.000} & \bf{1.000} & \bf{1.000} & \bf{1.000} & \bf{1.000} & \bf{1.000} & \bf{1.000} & \bf{1.000} & \bf{1.000} & \bf{0.999} & \bf{0.980} & \bf{0.858} & 0.525 & 0.172 \\ 
1760 &  & \bf{1.000} & \bf{1.000} & \bf{1.000} & \bf{1.000} & \bf{1.000} & \bf{1.000} & \bf{1.000} & \bf{1.000} & \bf{1.000} & \bf{1.000} & \bf{1.000} & \bf{1.000} & \bf{0.999} & \bf{0.985} & \bf{0.870} & 0.538 & 0.176 \\ 
1815 &  & \bf{1.000} & \bf{1.000} & \bf{1.000} & \bf{1.000} & \bf{1.000} & \bf{1.000} & \bf{1.000} & \bf{1.000} & \bf{1.000} & \bf{1.000} & \bf{1.000} & \bf{1.000} & \bf{0.999} & \bf{0.987} & \bf{0.879} & 0.550 & 0.179 \\ 
1870 &  & \bf{1.000} & \bf{1.000} & \bf{1.000} & \bf{1.000} & \bf{1.000} & \bf{1.000} & \bf{1.000} & \bf{1.000} & \bf{1.000} & \bf{1.000} & \bf{1.000} & \bf{1.000} & \bf{1.000} & \bf{0.988} & \bf{0.888} & 0.563 & 0.186 \\ 
1925 &  & \bf{1.000} & \bf{1.000} & \bf{1.000} & \bf{1.000} & \bf{1.000} & \bf{1.000} & \bf{1.000} & \bf{1.000} & \bf{1.000} & \bf{1.000} & \bf{1.000} & \bf{1.000} & \bf{1.000} & \bf{0.991} & \bf{0.897} & 0.576 & 0.188 \\ 
1980 &  & \bf{1.000} & \bf{1.000} & \bf{1.000} & \bf{1.000} & \bf{1.000} & \bf{1.000} & \bf{1.000} & \bf{1.000} & \bf{1.000} & \bf{1.000} & \bf{1.000} & \bf{1.000} & \bf{1.000} & \bf{0.992} & \bf{0.906} & 0.587 & 0.192 \\ 
2035 &  & \bf{1.000} & \bf{1.000} & \bf{1.000} & \bf{1.000} & \bf{1.000} & \bf{1.000} & \bf{1.000} & \bf{1.000} & \bf{1.000} & \bf{1.000} & \bf{1.000} & \bf{1.000} & \bf{1.000} & \bf{0.994} & \bf{0.914} & 0.600 & 0.196 \\ 
2090 &  & \bf{1.000} & \bf{1.000} & \bf{1.000} & \bf{1.000} & \bf{1.000} & \bf{1.000} & \bf{1.000} & \bf{1.000} & \bf{1.000} & \bf{1.000} & \bf{1.000} & \bf{1.000} & \bf{1.000} & \bf{0.994} & \bf{0.921} & 0.611 & 0.202 \\ 
2145 &  & \bf{1.000} & \bf{1.000} & \bf{1.000} & \bf{1.000} & \bf{1.000} & \bf{1.000} & \bf{1.000} & \bf{1.000} & \bf{1.000} & \bf{1.000} & \bf{1.000} & \bf{1.000} & \bf{1.000} & \bf{0.995} & \bf{0.926} & 0.623 & 0.208 \\ 
2200 &  & \bf{1.000} & \bf{1.000} & \bf{1.000} & \bf{1.000} & \bf{1.000} & \bf{1.000} & \bf{1.000} & \bf{1.000} & \bf{1.000} & \bf{1.000} & \bf{1.000} & \bf{1.000} & \bf{1.000} & \bf{0.996} & \bf{0.932} & 0.633 & 0.210 \\ 
2255 &  & \bf{1.000} & \bf{1.000} & \bf{1.000} & \bf{1.000} & \bf{1.000} & \bf{1.000} & \bf{1.000} & \bf{1.000} & \bf{1.000} & \bf{1.000} & \bf{1.000} & \bf{1.000} & \bf{1.000} & \bf{0.996} & \bf{0.938} & 0.642 & 0.214 \\ 
2310 &  & \bf{1.000} & \bf{1.000} & \bf{1.000} & \bf{1.000} & \bf{1.000} & \bf{1.000} & \bf{1.000} & \bf{1.000} & \bf{1.000} & \bf{1.000} & \bf{1.000} & \bf{1.000} & \bf{1.000} & \bf{0.997} & \bf{0.942} & 0.652 & 0.217 \\ 
2365 &  & \bf{1.000} & \bf{1.000} & \bf{1.000} & \bf{1.000} & \bf{1.000} & \bf{1.000} & \bf{1.000} & \bf{1.000} & \bf{1.000} & \bf{1.000} & \bf{1.000} & \bf{1.000} & \bf{1.000} & \bf{0.998} & \bf{0.948} & 0.664 & 0.221 \\ 
2420 &  & \bf{1.000} & \bf{1.000} & \bf{1.000} & \bf{1.000} & \bf{1.000} & \bf{1.000} & \bf{1.000} & \bf{1.000} & \bf{1.000} & \bf{1.000} & \bf{1.000} & \bf{1.000} & \bf{1.000} & \bf{0.998} & \bf{0.950} & 0.673 & 0.227 \\ 
2475 &  & \bf{1.000} & \bf{1.000} & \bf{1.000} & \bf{1.000} & \bf{1.000} & \bf{1.000} & \bf{1.000} & \bf{1.000} & \bf{1.000} & \bf{1.000} & \bf{1.000} & \bf{1.000} & \bf{1.000} & \bf{0.998} & \bf{0.955} & 0.684 & 0.229 \\ 
2530 &  & \bf{1.000} & \bf{1.000} & \bf{1.000} & \bf{1.000} & \bf{1.000} & \bf{1.000} & \bf{1.000} & \bf{1.000} & \bf{1.000} & \bf{1.000} & \bf{1.000} & \bf{1.000} & \bf{1.000} & \bf{0.999} & \bf{0.958} & 0.694 & 0.233 \\ 
2585 &  & \bf{1.000} & \bf{1.000} & \bf{1.000} & \bf{1.000} & \bf{1.000} & \bf{1.000} & \bf{1.000} & \bf{1.000} & \bf{1.000} & \bf{1.000} & \bf{1.000} & \bf{1.000} & \bf{1.000} & \bf{0.999} & \bf{0.963} & 0.701 & 0.239 \\ 
2640 &  & \bf{1.000} & \bf{1.000} & \bf{1.000} & \bf{1.000} & \bf{1.000} & \bf{1.000} & \bf{1.000} & \bf{1.000} & \bf{1.000} & \bf{1.000} & \bf{1.000} & \bf{1.000} & \bf{1.000} & \bf{0.999} & \bf{0.965} & 0.711 & 0.239 \\ 
2695 &  & \bf{1.000} & \bf{1.000} & \bf{1.000} & \bf{1.000} & \bf{1.000} & \bf{1.000} & \bf{1.000} & \bf{1.000} & \bf{1.000} & \bf{1.000} & \bf{1.000} & \bf{1.000} & \bf{1.000} & \bf{0.999} & \bf{0.969} & 0.719 & 0.243 \\ 
2750 &  & \bf{1.000} & \bf{1.000} & \bf{1.000} & \bf{1.000} & \bf{1.000} & \bf{1.000} & \bf{1.000} & \bf{1.000} & \bf{1.000} & \bf{1.000} & \bf{1.000} & \bf{1.000} & \bf{1.000} & \bf{0.999} & \bf{0.971} & 0.730 & 0.251 \\ 
2805 &  & \bf{1.000} & \bf{1.000} & \bf{1.000} & \bf{1.000} & \bf{1.000} & \bf{1.000} & \bf{1.000} & \bf{1.000} & \bf{1.000} & \bf{1.000} & \bf{1.000} & \bf{1.000} & \bf{1.000} & \bf{0.999} & \bf{0.973} & 0.738 & 0.254 \\ 
2860 &  & \bf{1.000} & \bf{1.000} & \bf{1.000} & \bf{1.000} & \bf{1.000} & \bf{1.000} & \bf{1.000} & \bf{1.000} & \bf{1.000} & \bf{1.000} & \bf{1.000} & \bf{1.000} & \bf{1.000} & \bf{1.000} & \bf{0.976} & 0.744 & 0.257 \\ 
2915 &  & \bf{1.000} & \bf{1.000} & \bf{1.000} & \bf{1.000} & \bf{1.000} & \bf{1.000} & \bf{1.000} & \bf{1.000} & \bf{1.000} & \bf{1.000} & \bf{1.000} & \bf{1.000} & \bf{1.000} & \bf{1.000} & \bf{0.977} & 0.753 & 0.263 \\ 
2970 &  & \bf{1.000} & \bf{1.000} & \bf{1.000} & \bf{1.000} & \bf{1.000} & \bf{1.000} & \bf{1.000} & \bf{1.000} & \bf{1.000} & \bf{1.000} & \bf{1.000} & \bf{1.000} & \bf{1.000} & \bf{1.000} & \bf{0.980} & 0.761 & 0.267 \\

\bottomrule

\end{tabular}

%% file: tbl-06-hassan-reeval.tex
\begin{tabular}{cc}
Segment Rating $+$ Document Context &
Document Rating $+$ Document Context \\[0.3em]
\begin{tabular}{clcccc}
\toprule
Ave. & Ave. z & $n$ & $N$ & System \\ 
\midrule
80.3 & 0.143\onesig & 902 & 1811 & REF-HT \\ 
76.6 & 0.038 & 904 & 1646 & REF-PE \\
76.5 & 0.036 & 863 & 1805 & Combo-6  \\
\bottomrule 
\end{tabular}
& 
\begin{tabular}{cccccc}
\toprule
Ave. & Ave. z & $n$ & $N$ & System \\ 
\midrule
78.9 & 0.184 & 114 & 216 & REF-HT \\ 
77.5 & 0.090 & 107 & 218 & REF-PE \\
76.0 & 0.050 & 106 & 238 & Combo-6 \\
\bottomrule 
\end{tabular}\\
\end{tabular}

%% file: sec-06-conc.tex
\section{Conclusion}

In this work, we explore issues relating to the reliability of machine translation evaluations. Firstly, we provide a detailed analysis of how translationese phenomena can adversely affect machine translation results. 
Our analysis of text that originated in a given language compared to that which had been created via human translation
showed that in general translated text is longer than text originally written in a given language.
Besides having different characteristics, in terms of the legitimacy of machine translation evaluation results, our analysis provides sufficient evidence that translationese is a problem for evaluation of systems, in particular in terms of comparison of system performance with automatic metrics such as BLEU.
This results in our first recommendation in future MT evaluations to \emph{avoid the use of test data that was created via human translation from another language}.

As described in Section \ref{relwork}, no previous work aiming to provide more certainty about conclusions of human parity in MT ticked all boxes. We therefore provided some missing analysis that should be included in the planning stage of future human evaluations of MT, particularly relevant to document-level evaluation that aims to investigate human-parity of MT. This analysis includes one of statistical power that will be useful as a reference for future MT evaluations to reduce the likelihood of future claims of human parity resulting from statistical ties produced from tests with low statistical power.

Finally, since evaluation of machine translation systems now involves several different criteria required to produce accurate and reliable results, we provide the following MT evaluation checklist for planning  upcoming MT evaluations:
\begin{enumerate}
   
    \item Test data creation direction -- reverse-created data should be avoided as it can potentially lead inaccurate results in particular in terms of BLEU scores;
    \item Human judge reliability -- either ensure high inter-annotator agreement levels or employ a method of human evaluation that has been shown to provide repeatable results such as Direct Assessment;
    \item Testing level (e.g. document or sentence) -- conclusions possible to be drawn from results are limited to the amount of context provided to human judges;
    \item Test language pairs -- only draw conclusions with reference to the tested language pairs;
    \item Test domain -- only draw conclusions with reference to the tested language domain;
    \item Translation sample size ($n$) -- numbers of distinct translations to be assessed should be planned prior to running a given evaluation to ensure sufficient statistical power (at least 80\%); $n$ should be reported and employed as the sample size for significance testing as opposed to the number of assessments;
    \item Human judgment sample size ($N$) -- numbers of assessments should also be reported;
    \item Meaningful overall statistic employed to distinguish performance of systems;
    \item Clustering via standard statistical significance testing.
\end{enumerate}

%% file: main.bbl
\begin{thebibliography}{19}
\expandafter\ifx\csname natexlab\endcsname\relax\def\natexlab#1{#1}\fi

\bibitem[{Baker et~al.(1993)Baker, Francis, and Tognini-Bonelli}]{Bakeretal:93}
Mona Baker, Gill Francis, and Elena Tognini-Bonelli. 1993.
\newblock Corpus linguistics and translation studies: Implications and
  applications.
\newblock In \emph{Text and Technology: In Honour of John Sinclair},
  Netherlands. John Benjamins Publishing Company.

\bibitem[{Bojar et~al.(2014)Bojar, Buck, Federmann, Haddow, Koehn, Leveling,
  Monz, Pecina, Post, Saint-Amand, Soricut, Specia, and Tamchyna}]{WMT:14}
Ondrej Bojar, Christian Buck, Christian Federmann, Barry Haddow, Philipp Koehn,
  Johannes Leveling, Christof Monz, Pavel Pecina, Matt Post, Herve Saint-Amand,
  Radu Soricut, Lucia Specia, and Ale\v{s} Tamchyna. 2014.
\newblock \href {http://www.aclweb.org/anthology/W/W14/W14-3302} {Findings of
  the 2014 workshop on statistical machine translation}.
\newblock In \emph{Proceedings of the Ninth Workshop on Statistical Machine
  Translation}, pages 12--58, Baltimore, Maryland, USA. Association for
  Computational Linguistics.

\bibitem[{Bojar et~al.(2013)Bojar, Buck, Callison-Burch, Federmann, Haddow,
  Koehn, Monz, Post, Soricut, and Specia}]{WMT:13}
Ond\v{r}ej Bojar, Christian Buck, Chris Callison-Burch, Christian Federmann,
  Barry Haddow, Philipp Koehn, Christof Monz, Matt Post, Radu Soricut, and
  Lucia Specia. 2013.
\newblock \href {http://www.aclweb.org/anthology/W13-2201} {Findings of the
  2013 {Workshop on Statistical Machine Translation}}.
\newblock In \emph{Proceedings of the Eighth Workshop on Statistical Machine
  Translation}, pages 1--44, Sofia, Bulgaria. Association for Computational
  Linguistics.

\bibitem[{Bojar et~al.(2016)Bojar, Chatterjee, Federmann, Graham, Haddow, Huck,
  Jimeno~Yepes, Koehn, Logacheva, Monz, Negri, Neveol, Neves, Popel, Post,
  Rubino, Scarton, Specia, Turchi, Verspoor, and Zampieri}]{WMT:16}
Ond\v{r}ej Bojar, Rajen Chatterjee, Christian Federmann, Yvette Graham, Barry
  Haddow, Matthias Huck, Antonio Jimeno~Yepes, Philipp Koehn, Varvara
  Logacheva, Christof Monz, Matteo Negri, Aurelie Neveol, Mariana Neves, Martin
  Popel, Matt Post, Raphael Rubino, Carolina Scarton, Lucia Specia, Marco
  Turchi, Karin Verspoor, and Marcos Zampieri. 2016.
\newblock \href {http://www.aclweb.org/anthology/W/W16/W16-2301} {Findings of
  the 2016 conference on machine translation}.
\newblock In \emph{Proceedings of the First Conference on Machine Translation},
  pages 131--198, Berlin, Germany. Association for Computational Linguistics.

\bibitem[{Bojar et~al.(2015)Bojar, Chatterjee, Federmann, Haddow, Huck, Hokamp,
  Koehn, Logacheva, Monz, Negri, Post, Scarton, Specia, and Turchi}]{WMT:15}
Ond\v{r}ej Bojar, Rajen Chatterjee, Christian Federmann, Barry Haddow, Matthias
  Huck, Chris Hokamp, Philipp Koehn, Varvara Logacheva, Christof Monz, Matteo
  Negri, Matt Post, Carolina Scarton, Lucia Specia, and Marco Turchi. 2015.
\newblock \href {http://aclweb.org/anthology/W15-3001} {Findings of the 2015
  workshop on statistical machine translation}.
\newblock In \emph{Proceedings of the Tenth Workshop on Statistical Machine
  Translation}, pages 1--46, Lisbon, Portugal. Association for Computational
  Linguistics.

\bibitem[{Bojar et~al.(2018)Bojar, Federmann, Fishel, Graham, Haddow, Huck,
  Koehn, and Monz}]{wmt18}
Ond\v{r}ej Bojar, Christian Federmann, Mark Fishel, Yvette Graham, Barry
  Haddow, Matthias Huck, Philipp Koehn, and Christof Monz. 2018.
\newblock \href {http://www.aclweb.org/anthology/W18-6401} {Findings of the
  2018 conference on machine translation (wmt18)}.
\newblock In \emph{Proceedings of the Third Conference on Machine Translation,
  Volume 2: Shared Task Papers}, pages 272--307, Belgium, Brussels. Association
  for Computational Linguistics.

\bibitem[{Callison-Burch et~al.(2007)Callison-Burch, Fordyce, Koehn, Monz, and
  Schroeder}]{WMT:07}
Chris Callison-Burch, Cameron Fordyce, Philipp Koehn, Christof Monz, and Josh
  Schroeder. 2007.
\newblock \href {http://www.aclweb.org/anthology/W/W07/W07-0218} {(meta-)
  evaluation of machine translation}.
\newblock In \emph{Proceedings of the Second Workshop on Statistical Machine
  Translation}, pages 136--158, Prague, Czech Republic. Association for
  Computational Linguistics.

\bibitem[{Callison-Burch et~al.(2008)Callison-Burch, Fordyce, Koehn, Monz, and
  Schroeder}]{WMT:08}
Chris Callison-Burch, Cameron Fordyce, Philipp Koehn, Christof Monz, and Josh
  Schroeder. 2008.
\newblock \href {http://www.aclweb.org/anthology/W/W08/W08-0309} {Further
  meta-evaluation of machine translation}.
\newblock In \emph{Proceedings of the Third Workshop on Statistical Machine
  Translation}, pages 70--106, Columbus, Ohio. Association for Computational
  Linguistics.

\bibitem[{Callison-Burch et~al.(2010)Callison-Burch, Koehn, Monz, Peterson,
  Przybocki, and Zaidan}]{WMT:10}
Chris Callison-Burch, Philipp Koehn, Christof Monz, Kay Peterson, Mark
  Przybocki, and Omar Zaidan. 2010.
\newblock \href {http://www.aclweb.org/anthology/W10-1703} {Findings of the
  2010 joint workshop on statistical machine translation and metrics for
  machine translation}.
\newblock In \emph{Proceedings of the Joint Fifth Workshop on Statistical
  Machine Translation and MetricsMATR}, pages 17--53, Uppsala, Sweden.
  Association for Computational Linguistics.
\newblock Revised August 2010.

\bibitem[{Callison-Burch et~al.(2012)Callison-Burch, Koehn, Monz, Post,
  Soricut, and Specia}]{WMT:12}
Chris Callison-Burch, Philipp Koehn, Christof Monz, Matt Post, Radu Soricut,
  and Lucia Specia. 2012.
\newblock \href {http://www.aclweb.org/anthology/W12-3102} {Findings of the
  2012 workshop on statistical machine translation}.
\newblock In \emph{Proceedings of the Seventh Workshop on Statistical Machine
  Translation}, pages 10--51, Montr{\'e}al, Canada. Association for
  Computational Linguistics.

\bibitem[{Callison-Burch et~al.(2009)Callison-Burch, Koehn, Monz, and
  Schroeder}]{WMT:09}
Chris Callison-Burch, Philipp Koehn, Christof Monz, and Josh Schroeder. 2009.
\newblock \href {http://www.aclweb.org/anthology/W/W09/W09-0401} {Findings of
  the 2009 {W}orkshop on {S}tatistical {M}achine {T}ranslation}.
\newblock In \emph{Proceedings of the Fourth Workshop on Statistical Machine
  Translation}, pages 1--28, Athens, Greece. Association for Computational
  Linguistics.

\bibitem[{Callison-Burch et~al.(2011)Callison-Burch, Koehn, Monz, and
  Zaidan}]{WMT:11}
Chris Callison-Burch, Philipp Koehn, Christof Monz, and Omar Zaidan. 2011.
\newblock \href {http://www.aclweb.org/anthology/W11-2103} {Findings of the
  2011 workshop on statistical machine translation}.
\newblock In \emph{Proceedings of the Sixth Workshop on Statistical Machine
  Translation}, pages 22--64, Edinburgh, Scotland. Association for
  Computational Linguistics.

\bibitem[{Cohen(1988)}]{cohen88}
Jacob Cohen. 1988.
\newblock \emph{Statistical power analysis for the social sciences}.
\newblock Hillsdale, NJ: Erlbaum.

\bibitem[{Graham et~al.(2016)Graham, Baldwin, Moffat, and
  Zobel}]{Grahametal:16}
Yvette Graham, Timothy Baldwin, Alistair Moffat, and Justin Zobel. 2016.
\newblock \href {https://doi.org/10.1017/S1351324915000339} {Can machine
  translation systems be evaluated by the crowd alone}.
\newblock \emph{Natural Language Engineering}, FirstView:1--28.

\bibitem[{Hassan et~al.(2018)Hassan, Aue, Chen, Chowdhary, Clark, Federmann,
  Huang, Junczys{-}Dowmunt, Lewis, Li, Liu, Liu, Luo, Menezes, Qin, Seide, Tan,
  Tian, Wu, Wu, Xia, Zhang, Zhang, and Zhou}]{hassanetal:18}
Hany Hassan, Anthony Aue, Chang Chen, Vishal Chowdhary, Jonathan Clark,
  Christian Federmann, Xuedong Huang, Marcin Junczys{-}Dowmunt, William Lewis,
  Mu~Li, Shujie Liu, Tie{-}Yan Liu, Renqian Luo, Arul Menezes, Tao Qin, Frank
  Seide, Xu~Tan, Fei Tian, Lijun Wu, Shuangzhi Wu, Yingce Xia, Dongdong Zhang,
  Zhirui Zhang, and Ming Zhou. 2018.
\newblock \href {http://arxiv.org/abs/1803.05567} {Achieving human parity on
  automatic chinese to english news translation}.
\newblock \emph{CoRR}, abs/1803.05567.

\bibitem[{Lambersky et~al.(2012)Lambersky, Ordan, and Wintner}]{lamb2012trans}
Gennadi Lambersky, Noam Ordan, and Shuly Wintner. 2012.
\newblock Language models for machine translation: Original vs. translated
  texts.
\newblock \emph{Computational Linguistics}, 38:4.

\bibitem[{L{\"a}ubli et~al.(2018)L{\"a}ubli, Sennrich, and
  Volk}]{laeublietal:18}
Samuel L{\"a}ubli, Rico Sennrich, and Martin Volk. 2018.
\newblock \href {http://arxiv.org/abs/1808.07048} {{Has Neural Machine
  Translation Achieved Human Parity? A Case for Document-level Evaluation}}.
\newblock In \emph{{EMNLP 2018}}, Brussels, Belgium. Association for
  Computational Linguistics.

\bibitem[{Papineni et~al.(2002)Papineni, Roukos, Ward, and Zhu}]{Papineni:2002}
Kishore Papineni, Salim Roukos, Todd Ward, and Wei-Jing Zhu. 2002.
\newblock {BLEU: A Method for Automatic Evaluation of Machine Translation}.
\newblock In \emph{Proceedings of the 40th Annual Meeting on Association for
  Computational Linguistics}, ACL '02, pages 311--318, Philadelphia,
  Pennsylvania.

\bibitem[{Toral et~al.(2018)Toral, Castilho, Hu, and Way}]{Toraletal:18}
Antonio Toral, Sheila Castilho, Ke~Hu, and Andy Way. 2018.
\newblock \href {https://arxiv.org/pdf/1808.10432.pdf} {Attaining the
  unattainable? reassessing claims of human parity in neural machine
  translation}.
\newblock \emph{CoRR}, abs/1808.10432.

\end{thebibliography}
